\documentclass[sn-apa]{sn-jnl}

\jyear{2021}%

\theoremstyle{thmstyleone}%
\theoremstyle{thmstyletwo}%

\theoremstyle{thmstylethree}%
\usepackage{lineno}

\begin{document}
\title[Deep Forest for PM2.5 Mapping from MAIAC AOD]{Using Deep Ensemble Forest for High Resolution Mapping of PM2.5 from MODIS MAIAC AOD in Tehran, Iran}


\author*[]{\fnm{Hossein} \sur{Bagheri}}\email{h.bagheri@cet.ui.ac.ir}



\affil[]{\orgdiv{Faculty of Civil Engineering and Transportation}, \orgname{University of Isfahan}, \orgaddress{\street{Azadi square}, \city{Isfahan}, \postcode{8174673441}, \country{Iran}}}




\abstract{\textcolor{red}{This is the pre-acceptance version, to read the final version, please go to Environmental Monitoring and Assessment on Springer, URL: \url{https://rdcu.be/c5ipl}}. High resolution mapping of PM2.5 concentration over Tehran city is challenging because of the complicated behavior of numerous sources of pollution and the insufficient number of ground air quality monitoring stations. Alternatively, high resolution satellite Aerosol Optical Depth (AOD) data can be employed for high resolution mapping of PM2.5. For this purpose, different data-driven methods have been used in the literature. Recently, deep learning methods have demonstrated their ability to estimate PM2.5 from AOD data. However, these methods have several weaknesses in solving the problem of estimating PM2.5 from satellite AOD data. In this paper, the potential of the deep ensemble forest method for estimating the PM2.5 concentration from AOD data was evaluated. The results showed that the deep ensemble forest method with $R^{2} = 0.74$ gives a higher accuracy of PM2.5 estimation than deep learning methods ($R^{2} = 0.67$) as well as classic data-driven methods such as random forest ($R^{2} = 0.68$). Additionally, the estimated values of PM2.5 using the deep ensemble forest algorithm were used along with ground data to generate a high resolution map of PM2.5. Evaluation of produced PM2.5 map revealed the good performance of the deep ensemble forest for modeling the variation of PM2.5 in the city of Tehran.}




\maketitle

\section{Introduction}\label{sec.intro}
Particulate matter (PM) is one of the main reasons for various diseases and mortality, particularly in industrialized and urban areas. Urbanization has increased the level of PM in recent decades and has caused severe health problems, which is one of the hot topics of environmental investigations and health studies. Studies illustrate particulate matters with sizes less than 2.5, called PM2.5, is the most harmful and dangerous pollutant exposing populations living in polluted areas such as metropolitan and industrialized areas \cite{lippmann2000association, klemm2000daily}. Consequently, monitoring the level of PM2.5 is a crucial and significant task for health surveillance and epidemiological studies.

Conventionally, PM2.5 concentration is measured using a network of air quality monitoring stations located in residential areas. The in-situ measurements are used to determine the air quality index of the covered area. Due to few monitoring stations, the level of pollutants cannot be estimated accurately in the interspaces of monitoring stations.

One of the main useful sources that can compensate for the lack of in-situ measurements is data collected by satellites which provide high resolution measurements with extensive coverage. Satellite sensors measure levels of pollutant particles (namely aerosol) in a column from the ground surface toward the sensor at the top of the atmosphere. In other words, the AOD product shows the amount of aerosols in the atmosphere. A higher value of AOD means a higher amount of aerosols in the atmosphere. Several satellite sensors such as Aqua and Terra have been designed to measure AOD. Both of the aforementioned sensors are mounted on the Moderate Resolution Imaging Spectroradiometer (MODIS) satellite, which has provided the possibility of daily measurement of aerosol particles in a wide area. The observations collected by Aqua and Terra sensors are converted into AOD data using different algorithms. Dark Target (DT) and Deep Blue (DB) algorithms are two well-known algorithms for AOD retrieval from satellite observations such as those acquired by Aqua and Terra. The DT algorithm is mostly used to retrieve AOD in areas with high vegetation density, which reduces its efficiency in urban areas (areas with a lack of dense vegetation). In contrast, DB is designed to retrieve AOD on bright surfaces in urban areas \cite{sayeretal, sayer2015effect}. However, both algorithms lead to AOD products with a resolution of 10 or 3 $km$, which are not desirable to monitor the variation of PM2.5 concentration in urban areas. Recently, the Multi-Angle Implementation of Atmospheric Correction (MAIAC) algorithm has been developed to retrieve AOD data from satellite observations, which has provided an opportunity to estimate AOD at a higher resolution (1 $km$) \cite{10.3389/frsen.2021.712093}. Several studies have demonstrated the potential of MAIAC AOD data for high resolution estimation of PM2.5 concentration \cite{di2016assessing, xiao2017full}. In order to retrieve AOD, this algorithm uses consecutive observations (time series) of Aqua and Terra sensors in a short period. As a result, it is possible to separate the reflection caused by dynamic phenomena such as aerosol and cloud versus the reflection caused by static surface phenomena.

The AOD data retrieved from satellite measurements can be potentially used for PM2.5 estimation \cite{christopher}. However, PM2.5 is a surface measurement, while AOD represents the content of aerosol in a column from the near-earth surface toward the sensor. Consequently, estimating the PM2.5 level from AOD is a challenging task.  
Different techniques have been applied for modeling and estimating PM2.5 from satellite AOD measurements which lie in three main categories: chemical simulations, statistical modeling, and semi-empirical modeling \cite{van2010global, SONG20141, LIN2015117}. Among these modeling techniques, statistical approaches are mainly based on data-driven methods (also called machine learning methods) and have been widely applied for modeling the AOD-PM2.5 relationship \cite{gupta_regression, ma2016satellite, YAO2018819, https://doi.org/10.1029/2004JD005025}.
In this regard, different machine learning techniques have been employed to estimate PM2.5 concentration \cite{Weizhen_2014, chen2020estimating, SUN2021144502, atmos9030105, https://doi.org/10.1029/2008JD011496, https://doi.org/10.1029/2008JD011497, https://doi.org/10.1002/2017GL075710, AHMAD2019117050}. 

Indeed, estimating PM2.5 from AOD measurements is considered a regression problem, and machine learning-based regression techniques try to explore a model based on the AOD variable as input and PM2.5 as output. The basic approach relies on a univariate regression model. Wang and Christopher showed a meaningful correlation between AOD and PM2.5 \cite{christopher}. However, the univariate model does not always have acceptable performance in constructing the relationship between those variables. The main reason is the complex behaviors of aerosols that are mostly affected by atmospheric conditions. Thus, some studies proposed to involve meteorological parameters such as boundary layer height, relative humidity, surface pressure, air temperature, etc., for modeling the relationship between PM2.5 and AOD \cite{gupta_regression, arciszewska2001importance}. Including more parameters in PM2.5 estimation from AOD demands the development of more efficient regression models. For this purpose, the univariate model was upgraded to a multivariate model \cite{gupta_regression}. In addition, several studies suggested more advanced models with the ability to employ more input features and explore sophisticated patterns \cite{SUN2021144502, https://doi.org/10.1002/2017GL075710}.


For PM2.5 estimation using AOD data, neural networks, a type of popular machine learning technology, have been used in several studies \cite{atmos9030105, https://doi.org/10.1029/2008JD011497}. Deep neural networks have demonstrated excellent performance in a variety of machine learning problems in recent years. Deep learning algorithms have also been used to estimate PM2.5 from satellite data and compared to decision tree ensemble methodologies. Several studies have demonstrated the superiority of deep learning over traditional machine learning models \cite{WANG2019128, rs12020264}. Li et al. designed a geo-intelligent network for PM2.5 estimate utilizing satellite AOD measurements based on deep belief networks \cite{https://doi.org/10.1002/2017GL075710}. For predicting PM2.5 and PM10 from AODs, Li employed autoencoder-based residual networks \cite{rs12020264}. The PM2.5-AOD modeling was discovered by Chen et al. using a self-adaptive deep neural network \cite{CHEN2021144724}. 

In general, deep neural networks suffer from some deficiencies such as having too many hyperparameters, demanding many training data, and requiring powerful computational facilities. Exploring patterns in sophisticated regression problems requires increasing complexity of network structure, i.e. adding more hidden layers. However, adding more layers makes the structure uninterruptible (black box). Moreover, increasing the network's capacity for providing sufficient complexity of the model will raise the probability of overfitting. On the other hand, training a deeper network is trickier and more complicated than a shallow neural network. Additionally, deep neural networks rely on weight optimization by backpropagation of errors. Backpropagation requires more computational loads and gradients could be exploded or vanished during training. 

Regarding the advantages and defects of deep neural networks, a recent development was established to devise a deep model such as a deep neural network but with non-differentiable modules (without backpropagation). For this, an ensemble approach based on decision trees estimators such as random forest, extremely randomized trees, etc., so-called deep ensemble forest, was developed. It takes the useful characteristics of deep neural networks such as layer-by-layer processing, in-model feature transformation, and sufficient model complexity while does not need any backpropagation process. In this way, the advantages of decision tree ensemble approaches and deep neural networks are combined in dealing with complex regression problems \cite{ijcai2017-497}.
\cite{10.3389/frsen.2021.712093,}
Considering the aforementioned properties of deep ensemble forest, the main objective of this paper is to assess the potential of deep ensemble forest for estimating PM2.5 concentration from MODIS AOD (1 $km$ MAIAC), weather data, and PM2.5 observations collected at air quality monitoring stations over Tehran city. 
The PM2.5 calculation using AOD data was performed in the research area using the 3 or 10 $km$ (DB or DT) products provided by MODIS. In a brief time of observations, Sotoudeheian and Arhami  calculated PM2.5 using 10 $km$ DT AODs. The correlation between anticipated and observed PM2.5 levels was around 0.55 \cite{sotoudeheian2014estimating}. Ghotbi et al. used 3 $km$ DT AOD and limited samples gathered from few ground stations for a very short time to estimate PM2.5 over Tehran with higher accuracy ($ R^{2} $ = 0.73) \cite{GHOTBI2016333}. Another study used the 10 $km$ MODIS AOD retrievals to estimate PM2.5 over Tehran. The achievements showed that machine learning improved accuracy by up to 80\% \cite{atmos10070373}.A further study in Tehran utilizing MAIAC AOD data for PM2.5 estimation found a correlation of smaller than 0.5, \cite{NABAVI2019889},  which is not ideal for high resolution AOD-based PM2.5 mapping. In a recent study, a map of PM2.5 at 1 $km$ resolution over the city was realized from MAIAC AOD products using a machine learning framework \cite{bagheri2019ASRJ}. 

As mentioned earlier, no study in the literature has focused on the capability of deep ensemble forest in comparison to deep neural networks for PM2.5 estimation from AOD data. Thus this paper also investigates the performance of deep ensemble forest in comparison to some common deep neural networks structures implemented in previous studies such as deep belief networks and deep autoencoders \cite{https://doi.org/10.1002/2017GL075710, rs12020264}.  In addition, the accuracy of deep ensemble forest in estimating PM2.5 is also compared with other advanced machine learning algorithms such as random forest, and extra trees, etc. that can outperform deep models in regression problems, particularly when dealing with tabular data and measurements.

This paper consists of seven sections. The introduction, the literature review, and the objectives of the paper were explained earlier in this section. The study area and datasets used in this investigation will be introduced in Section \ref{area}. The description of the deep ensemble forest algorithm is provided in Section \ref{method}. The implementation of deep ensemble forest for high resolution estimation of PM2.5 is realized through a framework including different stages, preprocessing of input materials; regression; and deployment which will be explained in Section \ref{setup}. Section \ref{setup} also provides experimental setups that were used for tuning and training the deep ensemble forest and other machine learning techniques. The results of deep ensemble forest regression are compared with other machine learning and deep learning techniques in the following (Section \ref{result}). Section \ref{map} discusses the performance of deep ensemble forest for 1 $km$ estimation of PM2.5 as daily and annual PM2.5 maps over Tehran. Finally, a conclusion of the investigation is presented in Section \ref{conclude}.

\section{Study Area and Datasets}\label{area}

\begin{figure}[t]
	
	\centering
	\centerline{\includegraphics[width=0.8\columnwidth]{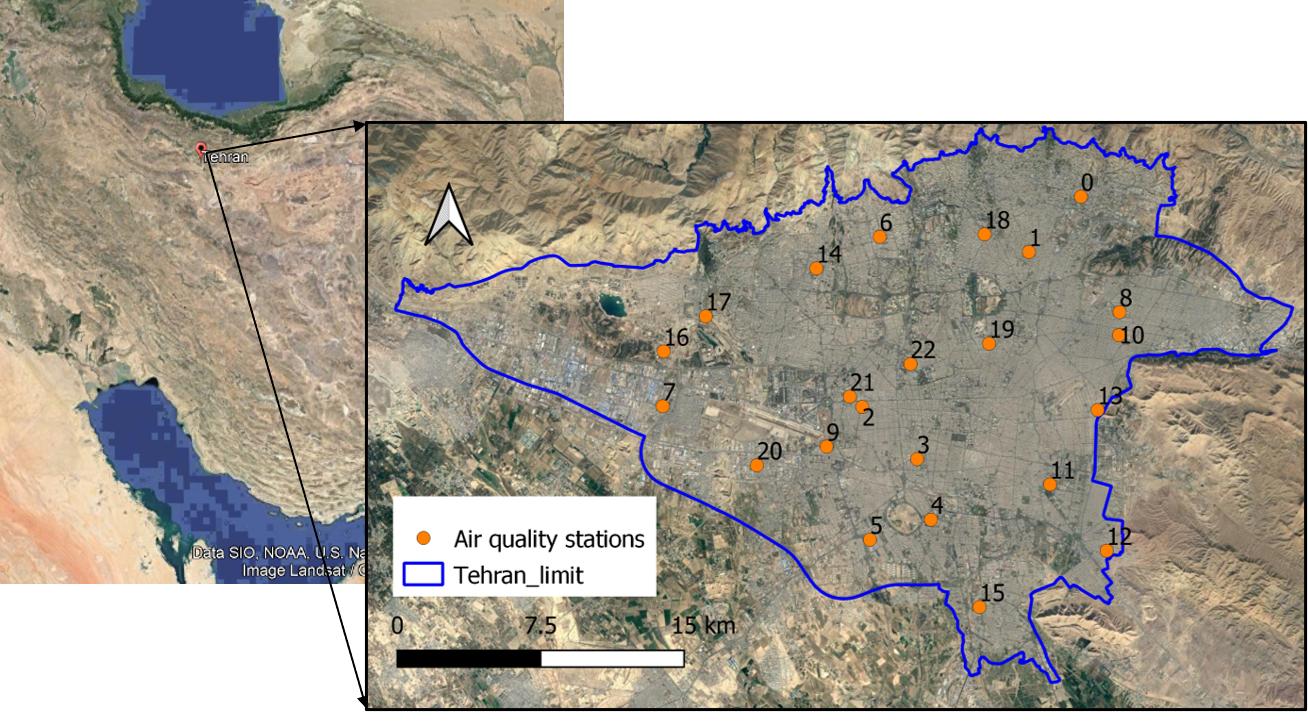}}
	\vspace{0.05cm}
	\caption{ A display of the study area. Numbers have been assigned by the author for further analysis in Section \ref{result}.}
	\label{study_area}
\end{figure}

The study region in this research is Tehran, Iran's capital, as a metropolitan residential area with a population of more than 16 million people. It is located at an area spread from latitude  35$ ^{\circ} $ 35$'$ N to  35$ ^{\circ}  $48$'$ N and longitude  51$ ^{\circ}  $17$'$ E to  51$ ^{\circ} $ 33$'$ E. Elevations change  between 900 m and 1800 m higher than the mean sea level. Tehran suffers from air pollution, particularly in the winter season, induced from different sources such as pollution emitted from mobile vehicles and emissions from manufactures located in the suburb.  Fig. \ref{study_area} displays the limit of Tehran city as well as locations of air quality stations. In-situ measurements are collected by these stations, which will be described more in the following. 

\subsection{PM2.5 In-situ Measurements}
As shown in Fig. \ref{study_area}, 23 air quality monitoring stations scattered across Tehran urban area have been launched by the Air Quality Control Company (AQCC: \url{http://airnow.tehran.ir/home/DataArchive.aspx}) of Tehran. PM2.5 is measured hourly at each station. However, measurements in some stations are missed occasionally due to technical problems. The daily PM2.5 is achieved by averaging hourly observations. In this study, daily measurements from Jan. 2013 to Dec. 2019 (7 years) were collected to model the PM2.5 variation.

\subsection{Satellite AOD Measurements}
Several types of AOD products are retrieved from observations by spaceborne sensors \cite{YAO2018819, https://doi.org/10.1029/2003GL018174, YOU20151156}. A famous product is the Multiangle Implementation of Atmospheric Correction (MAIAC) AOD, which is produced at 1 $km$ resolution from measurements generated by Aqua and Terra sensors on the MODIS platform. The MAIAC algorithm for AOD retrieval is achieved by time series processing of satellite observations, which allows dynamic phenomena like aerosols and clouds to be separated from static surface features. \cite{amt-11-5741-2018}. In this study, daily AOD products were generated from initial MAIAC AOD retrievals during the study period. The MODIS Version 6 MAIAC AOD data (MCD19A2.006) were achieved in HDF format from the EarthData portal of NASA (\url{https://search.earthdata.nasa.gov/search}). Finally, the AOD at each monitoring station was extracted by a 3 $\times$ 3 window from daily MAIAC grids. More details will be provided in Section \ref{frame}.

\subsection{Meteorological Data}\label{met}

Alongside the satellite AOD, meteorological data also have  significant roles in PM2.5 estimation \cite{atmos9030105, https://doi.org/10.1029/2008JD011496, https://doi.org/10.1029/2008JD011497, arciszewska2001importance}. For example,  boundary layer height has a vital role in distinguishing the surface PM2.5 level from total columnar aerosols \cite{gupta_regression}. In this study, different meteorological parameters such as temperature (T), dewpoint temperature (DT), planetary boundary layer height (PBLH), surface pressure (SP), wind properties (WS, WD), leaf area index (LAI), UV radiation (UV), and relative humidity (RH) were employed for PM2.5-AOD modeling. The meteorological data were derived from the ERA5 weather model that is developed by European Centre for Medium-Range Weather Forecasts (ECMWF) \cite{ER5, https://doi.org/10.1002/qj.3803}. The employed data can be achieved in NetCDF format from Copernicus Climate Data Store (\url{https://cds.climate.copernicus.eu/}).

In addition to the aforementioned features, other parameters such as positional data, i.e., latitude (Lat), and longitude (Long), and day of year (DOY) as a time variable were involved for final estimation.
The statistical summaries, and detailed descriptions of input features applied for PM2.5 modeling using satellite AOD have been collected in Tab. \ref{features_des}.

\begin{table}[ht!]
	\begin{center}

		\caption{List of input parameters employed for PM2.5 estimation in Tehran}
		\label{features_des}
		\begin{tabular*}{\textwidth}{l ccccc}
			Feature &Description & Max & Mean& Min & Std \\
			\toprule
	
			AOD &  Aerosol optical depth	 & 0.88 & 0.17 & 0.01 & 0.08 \\ 
			U & AOD Uncertainty & 1 & 0.41 & 0 & 0.24 \\\hline
			PM & PM2.5 & 64.80 & 28.99 & 0.60 & 12.02 \\\hline
			Lat & Latitude & 35.80 & 35.71 & 35.60 & 0.05  \\
			Long & Longitude	 &  51.51 & 51.38 & 51.24 & 0.08\\\hline
			T & 2m temperature	&  308.44 & 289.56 & 262.39 & 10.52 \\
			DT & 2m dewpoint temperature 	& 286.54 & 271.87 &250.52 & 5.22 \\
			PBLH & Planetary boundary layer height	& 1981.97 & 702.30 & 27.75 & 382.16 \\
			SP & Surface pressure	 &  2.33 & 1.81 & 1.30 & 0.27 \\ 
			LAI & Leaf area index	 &  0.53 & 0.50 & 0.47 & 0.01 \\ 
			WS & 10m wind speed	 & 4.91 & 1.81 & 0.70 & 0.47\\ 
			WD & 10m wind direction	 &  2.44 &-0.44 &-2.31 &0.61 \\
			UV & UV radiation	 &  139511.69 & 100547.60& 33686.45& 28484.86 \\ 
			RH & Relative humidity	  & 0.94 &0.69 &0.41 &0.12\\\hline
			DOY & Day of year	 & 365 & 185.69 & 1 & 96.42 \\	
			\hline		
		\end{tabular*}
\end{center}
\end{table}

\section{Methods} \label{method}
The main purpose of this study is to build a model (F) that can predict concentration of PM2.5 from input features, AOD; meteorological information; and other auxiliary data as below:

\begin{equation}\label{eq.1}
	\begin{split}
	PM = F(AOD, \ U, \ Lat, \ Long, \ T, \ DT, \ PBLH, \ SP, \ LAI, \ WS, \ WD, \\ \ UV, \ RH, \ DOY),
	\end{split}
\end{equation}
where $F$ is a functional model that associates AOD and other input features to PM2.5. The descriptions of other notations have been provided in Section  \ref{met} and Tab. \ref{features_des}.

The modeling process formulated by the equation above is a regression problem, and machine learning techniques can be applied to examine different hypotheses that better describe the relationship between input features and the target variable (PM2.5). As mentioned in Section \ref{sec.intro}, in the literature, different regression techniques based on decision tree ensemble approaches and deep neural networks have been applied \cite{SUN2021144502, https://doi.org/10.1002/2017GL075710}. In this study, deep ensemble forest as a state-of-the-art machine learning technique is developed for PM2.5 estimation and the potential of the developed method is evaluated to identify the best hypothesis that can describe the function $F$ mentioned in eq. \ref{eq.1}.

\subsection{Structure of Deep Ensemble Forest }
\begin{figure}[tb]
	
	\centering
	\centerline{\includegraphics[width=0.95\columnwidth]{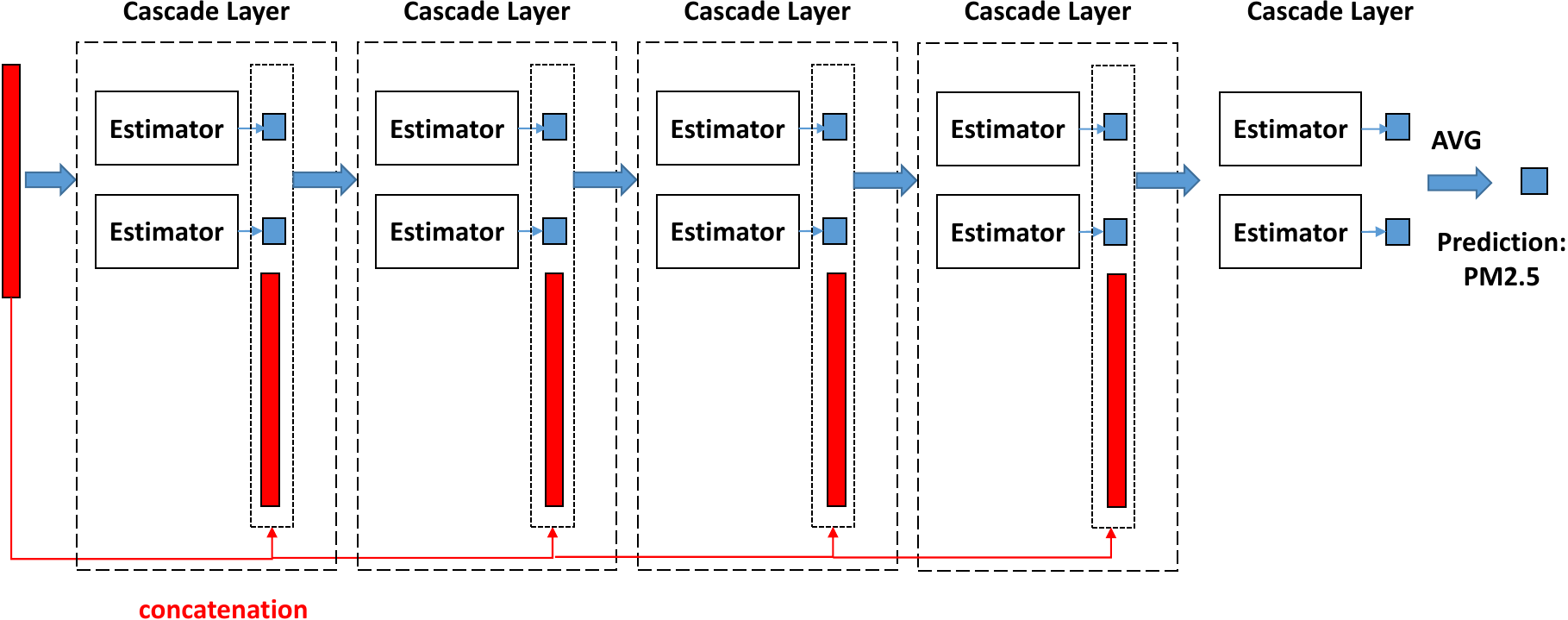}}
	\vspace{0.05cm}
	\caption{Deep ensemble forest structure (with four cascade layers including two estimators per layer) for PM2.5 estimation}
	\label{df}
\end{figure}

Deep ensemble forest is a deep model (similar to a deep neural network structure) using an ensemble of decision tree forests without the need for backpropagation \cite{ijcai2017-497}. In this way, its performance is highly competitive with deep neural networks while possessing fewer parameters and being more applicable with less amount of data. Fig. \ref{df} displays the structure of a deep ensemble forest regressor, which employs the cascade layers to generate a deep model. In each layer, estimators have the task of feature transformation and lead to developing new augmented features. In the following, the main modules constructing the structure of deep ensemble forest is explained in more detail.

\subsubsection{Cascade Layers}
Inspired by deep neural network structures that rely on layer-by-layer data processing and feature extraction, deep ensemble forest utilizes a cascade structure for representation learning. Each cascade layer receives features extracted by the preceding cascade layer and transforms the input features to generate new features that will be injected into the next layer. Each layer comprises \textit{n} number decision tree ensemble estimators that perform regression using the input features and target value (PM2.5 level). 
Each estimator generates a new feature that will be input into the next cascade layer. Consequently, a new \textit{n-dimensional} feature vector is achieved by \textit{n} estimators. This feature vector is also called an augmented feature vector and can effectively realize the representation learning for the features of the sample. However, before that, the produced new features by estimators are concatenated with primary features inputted into the cascade layer (represented as a red chunk in Fig. \ref{df}). In other words, the new feature vector includes transformed features generated by estimators within the layer as well as initial input features \cite{ijcai2017-497}. Mathematically, let $f^{inp}$ be the input feature vector, by using n estimators in a cascade layer j, the output feature vector can be formulated as below:

\begin{equation} \label{eq.2}
	\begin{split}
		L_{c}^{1} & = f^{inp} \cup_{i}^{n} \ R_{i}^{1} \ \cup_{i}^{n} E_{i}^{1} \\
		L_{c}^{j} & = L_{c}^{j-1} \cup_{i}^{n} \ R_{i}^{j} \cup_{i}^{n} E_{i}^{j},
	\end{split}
\end{equation}
where $L_{c}^{j}$ is the feature vector output from the jth cascade layer, $R_{i}^{j}$  , and $E_{i}^{j}$ are the output of random forest and extremely randomized trees estimators, respectably, embedded in the $j^{th}$ layer, and $\cup$ denotes the concatenation operator. Note that each decision tree in estimators randomly selects $\sqrt{d}$ feature from $L_{c}^{j-1}$.
The feature extraction and concatenation are continued until the last layer, in which the random forest and extremely randomized trees estimators predict the final outputs using feature values established by cascade layers. The outcome is an average of predictions achieved by estimators at the last layer.

\begin{figure}[tb]
	\centering
	\begin{minipage}[b]{0.8\linewidth} 
		\centering
		\centerline{\includegraphics[width=1\columnwidth]{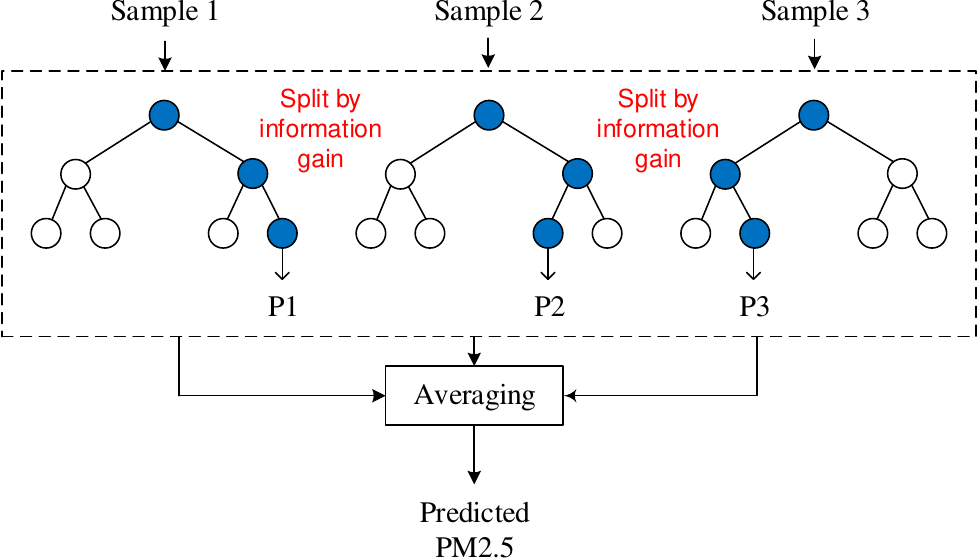}}
	\end{minipage}
	\vspace{0.05cm}
	\caption{Random Forest estimator for PM2.5 prediction}
	\label{rf}
\end{figure}

\begin{figure}[tb]
	\centering	
	\begin{minipage}[b]{0.8\linewidth} 
		\centering
		\centerline{\includegraphics[width=1\columnwidth]{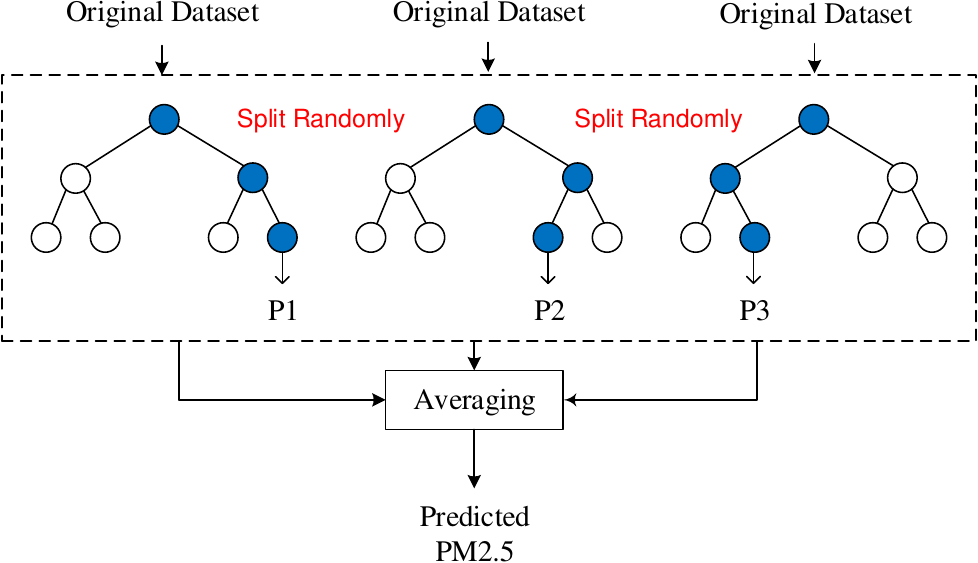}}
	\end{minipage}
	\caption{Extra Trees estimator for PM2.5 prediction}
	\label{et}
\end{figure}

\subsubsection{Estimators}
Two types of estimators, random forest regressor and extremely randomized trees formed based on the classification and regression tree (CART), can be employed at the heart of cascade layers to increase diversity as a crucial characteristic for ensemble learning. 
The estimators transform the input features to new augmented features, which will be concatenated with original features and be injected into the next layer. 

As shown in Fig. \ref{rf}, random forest as an ensemble method works based on fitting a decision tree on a bootstrapped dataset (samples) \cite{james2013introduction}. For this, a decision tree is built using a random subset of features. The split at each node is achieved by the optimal feature selected from the random subset using gain information. Building decision trees is repeated for $m$ times by a new subset of features and a new bootstrapped dataset. Finally, the random forest model is formed based on the ensemble of $m$ fitted decision tree regressors. 

Extremely randomized trees (also called extra trees) is similar to random forest; However, in extra trees, one more randomization step has been added in the process of building decision trees \cite{geurts2006extremely}. Instead of selecting the best feature for node split, the algorithm  chooses the feature completely random from the subset at node split (Fig. \ref{et}). In this way, an extremely randomized trees estimator adds more diversity to the deep ensemble forest and are computationally more efficient than random forest estimators. 

\subsection{Comparative Study}\label{comp_study}
In this study, the performance of the tuned deep ensemble forest for PM2.5 estimation was benchmarked in comparison with different machine learning algorithms. 
The simplest model can be established by using a simple linear regression technique, in which the AOD extracted at the location of the monitoring station is an input variable, and the ground PM2.5 measurement is an independent variable. The achieved linear equation is used for estimating PM2.5 in other locations. In addition to satellite retrieved AOD, meteorological data have shown their efficiency in estimating PM2.5 concentration as auxiliary input features. In this situation, multivariate linear regression can be applied for the modeling.

Ensemble approaches combine decision trees as weak learners to generate more efficient predictors by diminishing defects of decision trees such as overfitting. The performance of algorithms such as random forest and extra trees which were explained earlier will also be assessed for AOD-PM2.5 modeling. 
In recent years, deep neural networks have also been presented as an outperforming, state-of-the-art algorithm for many machine learning problems. 
The main characteristic of deep neural network is automatically feature extraction and engineering from raw input data such as images \cite{goodfellow2016deep}. Different types of deep neural networks have been developed for different problems. 
One type of deep neural network that can be developed for regression problems is deep autoencoder decoder (DAE) \cite{goodfellow2016deep}. In this type of network, an encoder extracts higher-level features from raw data. The network structure is restricted to a bottleneck at the midpoint of the model from which, the network attempts to reconstruct the input data. After fitting the model, the reconstruction part is discarded, and the model up to the bottleneck is applied for feature extraction. In the autoencoder part, input features are inputted into the network, and more higher-level learned features are generated through the network by the composition of lower-level features. This automatic feature extraction and engineering provide an opportunity to learn sophisticated function mapping between input features and target variables \cite{liu2017nirs}. In the PM2.5 concentration problem, the input features are injected into a DAE to generate more beneficial, higher-level features. Then the achieved features are injected into a regression method such as SVR  to  predict PM2.5 values observed at the monitoring stations, ultimately. This process is performed by training in which the network tries to extract more advanced features as well as their weights, participating in forming the regression model. 


The deep belief network (DBN), which is made up of a stack of restricted Boltzmann machines (RBM), is another deep learning structure that can be used to estimate PM2.5.
Each RBM comprises two main layers: the visible layer as the input layer and the hidden layer. On the forward pass, activations are produced using the hidden units, and then, the features are reconstructed by visible units on the backward pass using fine-tuning. This process is performed in each RBM layer with an output of the previous layer \cite{hinton2006fast}. The constructed features at the final RBM are converted into PM2.5 values. 


\section{Experimental Setup}\label{setup}

\subsection{Data Preparation for PM2.5 Estimation by Deep Ensemble Forest }\label{frame}
\begin{figure*}[t!]
	\begin{center}
		\includegraphics[width=1\textwidth]{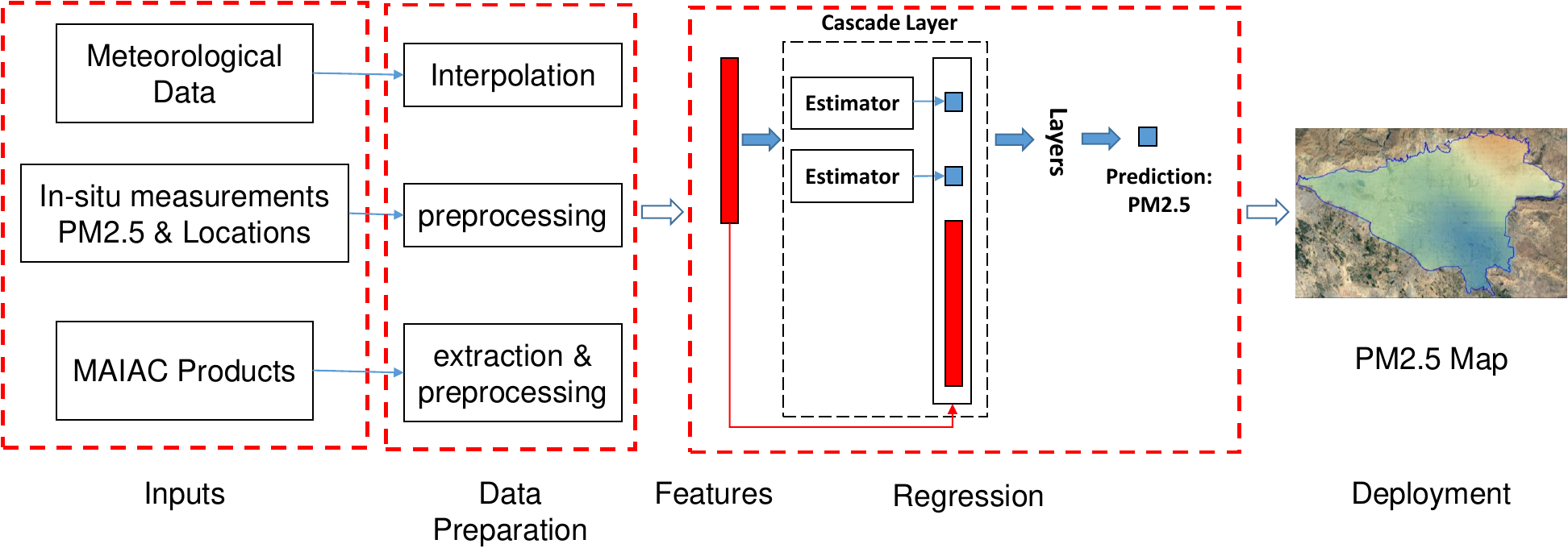}
		\caption{The framework designed for estimation of PM2.5 over Tehran at 1 $km$ resolution}
		\label{fig.frame}
	\end{center}
\end{figure*}

The main objective of this paper is to apply the tuned deep ensemble forest for generating high resolution maps of PM2.5 using MAIAC AOD data. 
Before developing any regression models such as deep ensemble forest and subsequently generating a high resolution PM2.5 map, it is required to prepare input features and corresponding target values (PM2.5) in a structured (tabular) format. A framework that is appropriate for the study area is created for this purpose. The suggested framework is depicted in detail in Fig. \ref{fig.frame}. Data preprocessing, regression, and deployment are the three key modules of the framework.
The data preprocessing includes several steps involving PM2.5 data preprocessing, AOD data extraction, and meteorological data preparation.   

At first, PM2.5 data collected by AQCC of Tehran at air quality monitoring stations are preprocessed. After heating the ambient air to 50$ ^{\circ} $C, TEOM measures PM2.5 values at stations, and the mass of dry PM is collected as measured PM2.5. Consequently, the reported values should be modified according to the relation below \cite{bagheri2019ASRJ}:

\begin{equation}
	PM_{c}= PM (1-\dfrac{RH}{100})^{-1}, \label{pm}
\end{equation}
where $ PM_{c} $ is the modified level of observed $ PM $ at the ground station and $RH$ is the relative humidity.



Daily PM2.5 values contain outliers that should be detected before usage. Outliers can simply be detected and subsequently removed by the interquartile range (IQR) \cite{yang2019outlier}. In this way, a PM2.5 observation that is between $Q_{1}-IQR$ and $Q_{3}+IQR$ is assumed an inlier. Note that $Q_{1}$ and $Q_{3}$ are the first and third quartiles of measured PM2.5, respectively.

After PM2.5 data preparation, MAIAC AOD data should be normalized. This is realized using PBLH delivered by ECMWF model. The formulation for this normalization is as below \cite{wang2010satellite}: 
\begin{equation}
	nAOD = \dfrac{AOD}{PBLH}, \label{naod2}
\end{equation} 

MAIAC AOD data are retrieved from observations collected by Aqua and Terra sensors on the MOIDS platform. Previous investigations have demonstrated that averaging of AODs provided by those sensors from observations of different orbital passes can be correlated with daily PM2.5 data \cite{lee2011novel, hu2014estimating}. However, the delivered MAIAC AOD retrievals can not be simply averaged to estimate PM2.5. As shown in Fig. \ref{fig.frame}, to accurately estimate PM2.5 from MAIAC AOD data, several preprocessing steps should be performed on them. In more detail, a MAIAC AOD product involves different AOD retrievals achieved from observations of two Aqua and Terra sensors in several orbital passes. Thus, the first step of preprocessing is to average retrievals of each sensor separately to produce a daily AOD retrieval. The outputs will be a daily mean of AOD for Aqua ($AOD_{A}$) as well as a daily mean of AOD for Terra ($AOD_{T}$).

In addition, for the study area, a linear correlation between the AODs of two sensors with a coefficient of determination of $0.72$ can be established. In this way, missing AOD values in each sensor can be estimated based on the available values in another sensor using linear regression. After filling in the missing values, the final AOD is achieved by merging the retrievals from both sensors (Aqua and Terra) by an averaging as below:

\begin{equation}
	AOD = \dfrac{AOD_{A}+AOD_{T}}{2}. \label{AODmean}
\end{equation}

After AOD data preparation, they are extracted at each location of the air quality station. For this purpose, a window with a size of 3$ \times $3 is applied at the location of each station to coincide with the corresponding MODIS pixel. The final AOD at the location is the mean of AOD values within the window. 

In order to extract only high quality AODs, several criteria are considered. First, the output AOD from the window should be an average of at least three pixels that have valid values (non missing). Second, the standard deviation of AOD values within the window should be less than 0.5. This criterion prevents engaging severely fluctuated AOD values at the neighborhood in final computation (AOD averaging). 

In parallel, the quality of AOD data is also extracted from the “Quality Assessment” (QA) file \cite{lyapustin2018modis}. According to recommendations presented in the user guide of MAIAC data, AOD data are filtered based on the adjacency mask and cloud mask in QA. The AODs have a normal condition or are clear in the adjacency mask as well as those that have been flagged clear or possibly cloudy in the cloud mask are selected. 

Filtering AODs by aforementioned conditions reduces the number of samples that can affect the accuracy of models, particularly deep neural networks. Instead of filtering, a new feature describing the uncertainty of AOD measurements  can be defined as below: 

\begin{equation}
	U = \dfrac{N_{b}}{N}, \label{uncertainty}
\end{equation} 
where, $U$ denotes the AOD uncertainty, $N_{b}$ denotes the number of pixels satisfies conditions mentioned above, and $N$ is the total number of pixels in a window. 
In other words, U identifies AOD uncertainty as the probability of the existence of AODs with the best quality in a window.  

In addition to AOD data, meteorological data are also employed as valuable features for estimating PM2.5. As mentioned in Section \ref{met}, the meteorological data derived from ECMWF model (ERA5) must be calculated at the position of a ground station. Thus, the value of each meteorological parameter was interpolated at the position of the ground station using the kriging approach. The settings for interpolation of each parameter, such as kriging type and semi-variogram type, were determined by a grid search strategy \cite{bagheri2014}.


After preparing data in a structured format, they can be inputted into the regression module but before that, it is needed to normalize input values. As shown in Tab. \ref{features_des}, the input features have diverse scales that can influence the performance of regression. Consequently, a min-max normalization strategy was applied to scale values of all input features between 0 and 1 as below: 

\begin{equation}
	f^{n} = \dfrac{f^{i}-f_{min}}{f_{max}-f_{min}}, \label{min-max}
\end{equation}
where  $ f^{i} $ is the input feature, $ f_{max} $ and $ f_{min} $ are minimum and maximum values of corresponding input feature, and $ f^{n} $ denotes the normalized version of input feature value.

After data preparation, a machine learning algorithm is tuned for estimating the PM2.5 from the selected input features. After tuning the algorithm using training data, the performance of the achieved model is evaluated by test data. The main focus of this study is to develop the deep ensemble forest algorithm for this regression task. Additionally, other regression techniques are also examined in comparison to deep ensemble forest for PM2.5 concentration estimation. More details of deep ensemble forest implementation and tuning and comparison with other regression models will be provided in Section \ref{DF_set}. 

As will be shown in Section \ref{result}, the developed deep ensemble forest algorithm will outperform other models, and as a result, it will be used for high resolution estimation of PM2.5 over the study area. In this phase, similar preprocessing procedures explained earlier are accomplished to prepare AOD measurements and meteorological data at locations (pixels) where PM2.5 has not been sensed by ground sensors. In the following, PM2.5 can be estimated at each pixel of MAIAC AOD products by using the fitted deep ensemble forest model. In other words, in the deployment step, most of the processes performed in preprocessing and regression modeling stages are fulfilled on target pixels. The output is a 1 $km$  PM2.5 map that represents the variation of PM2.5 over the study area.

\subsection{Deep Ensemble Forest Setup}\label{DF_set}

Deep ensemble forest regression was set up by a set of samples including input variables and corresponding PM2.5 through the k-fold cross-validation. The k-fold cross-validation was applied for hyperparameter setting as well as mitigating overfitting. In general, as k increases, the amount of bias in the estimation decreases. However, increasing the amount of k requires more computational cost, particularly, in case, the data volume is large. As a rule of thumb, k is considered to be 5 or 10 in most cases \cite{anguita2012k}. In this research, according to the amount of data and the available computing facilities (available RAM = 16 GB), k was configured as 5. From around 20k samples, 70\% of data was employed for k-fold cross-validation. The main hyperparameters of deep ensemble forest were set up using a grid search strategy with 5-fold cross-validation. The grid search found that the optimal number of cascade layers should be 2, the number of estimators should be 4 (two extra trees and two random forests), and the number of trees in each estimator should be 2000. Each tree was expanded until all leaves have 1 sample and no constraint was considered for trees.

After training, the performance of the developed deep ensemble forest was evaluated by test data (30\% percent of remaining data) using standard metrics such as root mean square error (RMSE), mean absolute error (MAE), coefficient of determination ($R^{2}$), and absolute percentage error (APE) formulated as below: 

\begin{equation}
	RMSE = \sqrt{\frac{\sum_{i=1}^n {(y_{i}-y_{i}^p})}{n}}, \label{RMSE}
\end{equation}

\begin{equation}
	MAE = \frac{ \sum_{i=1}^n \left\lvert{y_{i}-y_{i}^p}\right\rvert }{n}, \label{MAE}
\end{equation}

\begin{equation}
	R^{2} = \left[ \frac{n \sum_{i=1}^n {y_{i} y_{i}^p} - \sum_{i=1}^n {y_{i}} \sum_{i=1}^n {y_{i}^p}}	{\sqrt{\left[n \sum_{i=1}^n {y_{i}^2}-\left( \sum_{i=1}^n {y_{i}}\right)^2\right] \left[n \sum_{i=1}^n {\left( y_{i}^p\right) ^2}-\left( \sum_{i=1}^n {y_{i}^p}\right)^2\right]}}\right]^{2}, 
	\label{R2}
\end{equation}

\begin{equation}
	APE = \frac{\sum_{i=1}^n \left\lvert {y_{i}-y_{i}^p} \right\rvert }{\sum_{i=1}^n {y_{i}}}, \label{APE}
\end{equation}

In equations above, $ y_{i} $ is in-situ PM2.5 measurement, $ y_{i}^p $  is predicted PM2.5 by the regression model, and $ n $ denotes the number of samples. 

	%
	%

\subsection{Setup Algorithms to Compare with Deep Ensemble Forest}
In this study, the performance of the tuned deep ensemble forest for PM2.5 estimation was benchmarked in comparison with different machine learning algorithms such as multivariate regressor, random forest, and extremely randomized trees which were described in Section \ref{comp_study}. Similar to the deep ensemble forest setup, hyperparameters of each algorithm were determined using grid search strategy with 5-fold cross-validation as: multivariate regressor (no hyperparameters); random forest (number of trees is 500, maximum depth is 10; maximum features is 0.5, minimum samples in a leaf is 1); extra trees (number of trees is 1000, maximum depth is 10, maximum features is 0.8, minimum samples in a leaf is 1). All aforementioned algorithms were implemented using scikit-learn library \cite{sklearn_api} in Python.

In addition, two deep learning structures were implemented by the author in TensorFlow 2 to comparatively study the performance of deep ensemble forest. For this aim, a deep belief network (DBN) based on the restricted Boltzmann machine (RBM) was carried out. The optimal structure of the network was achieved by validation data. In this research, a DBN consisting of two RBM blocks, including one visible layer (input layer) and two hidden layers within 64, 10 neurons, was trained for PM2.5 estimation. Other hyperparameters were set to be: the learning rate of RBM is 0.01, the learning rate of the network is 0.001, the optimizer is stochastic gradient descent,  number of epochs for training RBM is 50, number of backpropagation iteration is 200, mini-batch is 256, the activation function is ReLU, and loss is MSE. 

\begin{figure}[tb]
	
	\centering
	\centerline{\includegraphics[width=0.95\columnwidth]{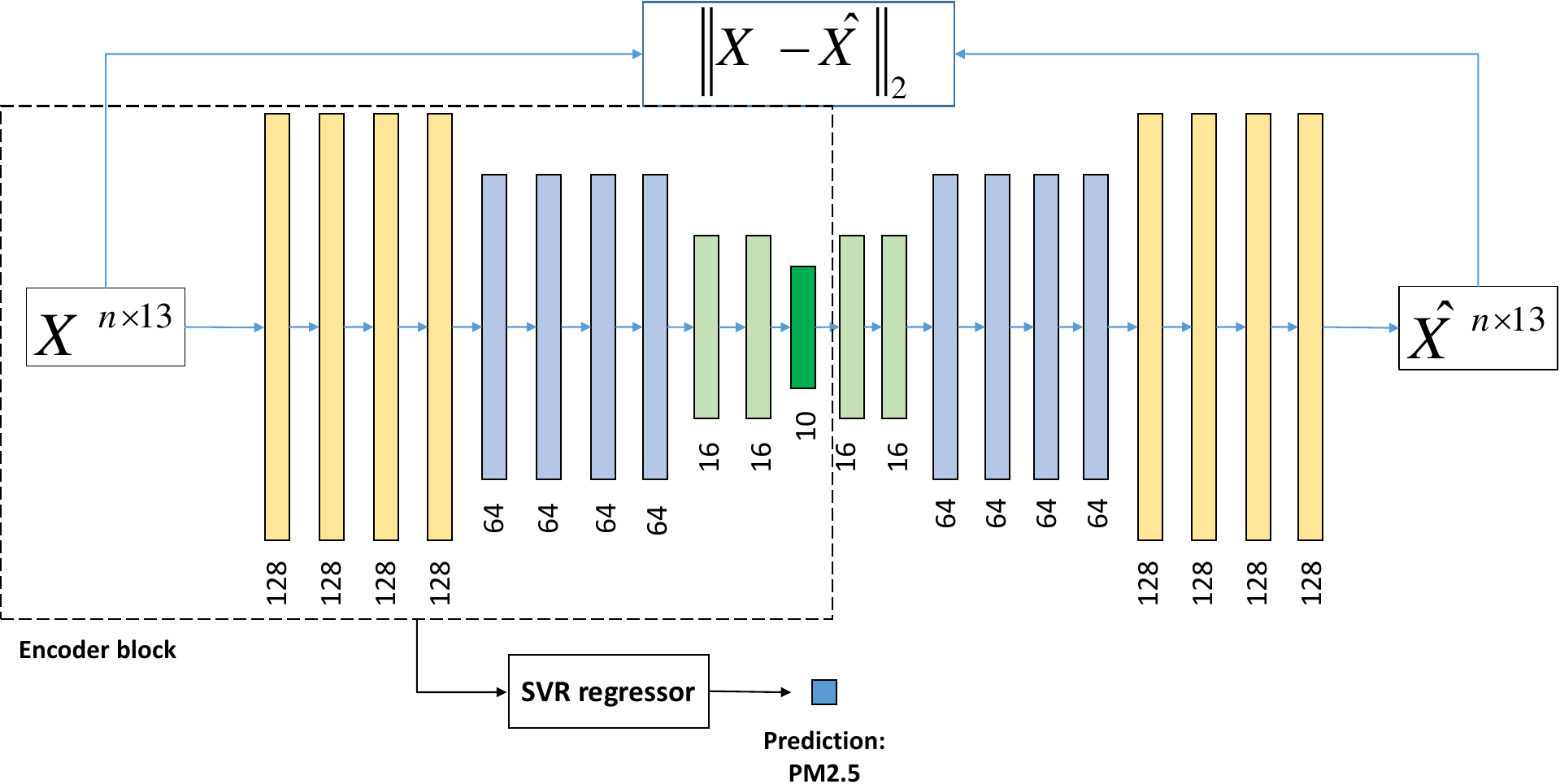}}
	\vspace{0.05cm}
	\caption{The structure of deep autoencoder-decoder used for feature extraction. The SVR regressor applies the generated features for PM2.5 estimation.}
	\label{dl}
\end{figure}

The second experiment was accomplished by a support vector regressor (SVR) using features extracted from a deep autoencoder (DAE). For this, a deep autoencoder-decoder was employed to extract new transformed features. The structure of applied deep autoencoder-decoder is shown in Fig \ref{dl}. 

The number of neurons in each layer is expressed in the diagram of the network illustrated in Fig. \ref{dl}. The features derived from the bottleneck layer (colored by dark green) are imported as input transformed variables into a SVR regressor. Other hyperparameters were set as: DAE (optimization is Adam, learning rate is 0.001, activation function is ReLU, loss function is MSE, regularization term is formed based on $L_{2}$ norm using weight penalty of 0.001, epoch is set 200, mini-batch size is 256) + SVR (kernel is chosen radial basis function, regularization parameter is 100, and epsilon is 0.1). 

In all experiments, the same test dataset was employed for comparing results. 

\section{Results and Discussion}\label{result}
\subsection{Performance of Deep Ensemble Forest }\label{DFperf}

\begin{figure}[tb]
	\begin{minipage}[b]{0.47\linewidth} 
		\centering
		\centerline{\includegraphics[width=1\columnwidth]{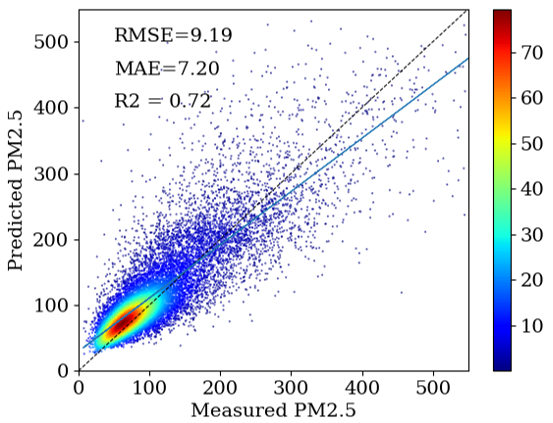}}
		\centerline{(a) Train Performance}\medskip
	\end{minipage}
	\begin{minipage}[b]{0.47\linewidth} 
		\centering
		\centerline{\includegraphics[width=1\columnwidth]{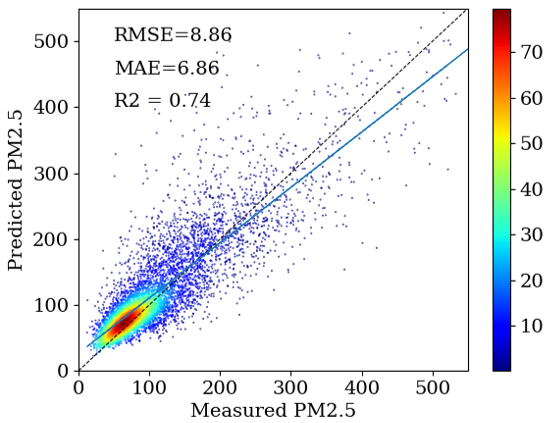}}
		\centerline{(b) Test Performance}\medskip 
	\end{minipage}
	\caption{Performance of the developed deep ensemble forest; a) on training data by k-fold cross-validation, and b) on test data}
	\label{perfomance}
\end{figure}

Fig. \ref{perfomance} illustrates the performance of the developed deep ensemble forest for building the PM2.5-AOD relationship. The correlation plots are drawn for both training and test datasets. As shown in Fig. \ref{perfomance}, the correlation between predicted PM2.5 and measured PM2.5 (at ground stations) is 0.72 and RMSE and MAE are 9.19 $ \frac{\mu g}{m^{3}} $ and 7.20 $ \frac{\mu g}{m^{3}} $, respectively. The results imply that the hyperparameter determination has been accomplished correctly by train data through the k-fold cross validation. However, it is required to evaluate the performance of the model with test data. In this regard, the performance of the algorithm was measured by test data. The correlation, RMSE, and MAE of the deep ensemble forest on test data were 0.74, 8.86 $ \frac{\mu g}{m^{3}} $, 6.86 $ \frac{\mu g}{m^{3}} $. It should be noted that the correlation plots were depicted based on scaled values of predicted PM2.5 (corrected PM2.5 i.e. $ PM_{c} $, See eq. \ref{pm}). However, the RMSE and MAE values were calculated after rescaling values to the initial scale. 

\begin{table*}[tb]
	\centering \footnotesize
	\caption{The performances of deep ensemble forest in comparison to other machine learning models applied for PM2.5 estimation}
	\label{Tab.result}
	\begin{tabular}{l cccc cccc}
		
		& \multicolumn{4}{c}{Model training: k-fold cross-validation}& \multicolumn{4}{c}{Model testing} \\
		Method & RMSE  $\frac{ \mu  g}{ m^{3}}$& MAE  $\frac{ \mu  g}{ m^{3}}$ & $ R^{2} $ & APE  & RMSE  $\frac{ \mu  g}{ m^{3}}$ & MAE $\frac{ \mu  g}{ m^{3}}$ & $ R^{2} $ & APE \\
		\toprule
		
		Multivariate &	13.45& 10.42&	0.54&	36\% & 12.55&	9.94 &	0.59 & 34\% \\
		
		Random Forest&	9.81&	7.83&	0.68&	27\% &9.64& 7.64& 0.69& 26\% \\
		Extra Trees&	9.83&	7.88&	0.67&	27\% &9.77&	7.79&	0.68 & 27\% \\
		
		DBN &	10.16&	7.94&	0.66& 27\%&	9.69&	7.40&	0.68 & 26\% \\
		DAE + SVR&	10.15 & 7.75&	0.67&	27\%& 9.94&	7.47&	0.67& 26\% \\
		deep ensemble forest&	\textbf{9.19} &	\textbf{7.20}&	\textbf{0.72}&	\textbf{25\%}& \textbf{8.86}&	\textbf{6.86}&	\textbf{0.74}& \textbf{24\%} \\
		
	\end{tabular} 	
\end{table*}

Tab. \ref{Tab.result} collects the performance of other machine learning techniques in comparison to the deep ensemble forest model on both training and test phases. This benchmarking gives a fair judgment of the ability of versatile machine learning techniques, deep learning, and deep ensemble forest for PM2.5 estimation. The comparison demonstrated that the deep ensemble forest was ranked first in terms of performance measured by different metrics either on the train or test data. After that, based on MAE metric, the best results were obtained by DBN and DAE employing SVR as deep learning techniques owned to layer-by-layer processing and feature extraction by a deep structure, while the performance of DBN is slightly better than DAE. Compared to different traditional machine learning techniques, the highest performance was obtained by random forest as a decision tree ensemble approach. According to RMSE and $ R^{2} $, random forest also gives the highest accuracy among other comparative algorithms.
The lowest accuracy was achieved by multivariate linear regression because of the simplicity of the method that could not model sophisticated patterns existing between input features and the output predictable variable (PM2.5) properly.

In general, DBN, DAE+SVR algorithms, and random forest have almost similar performances based on different metrics (See Tab. \ref{Tab.result}). The main difference between three algorithms was represented by the MAE metric on both train and test data. The differences in MAE values are originated from the existing noise in input feature values. The AE structure reduced the noise effect own to its denoising property, while, the  performance of random forest was more affected by noise that has led to slightly worse MAE. 

\begin{figure}[H]
	\begin{minipage}[b]{1\linewidth} 
		\centering
		\centerline{\includegraphics[width=1\columnwidth]{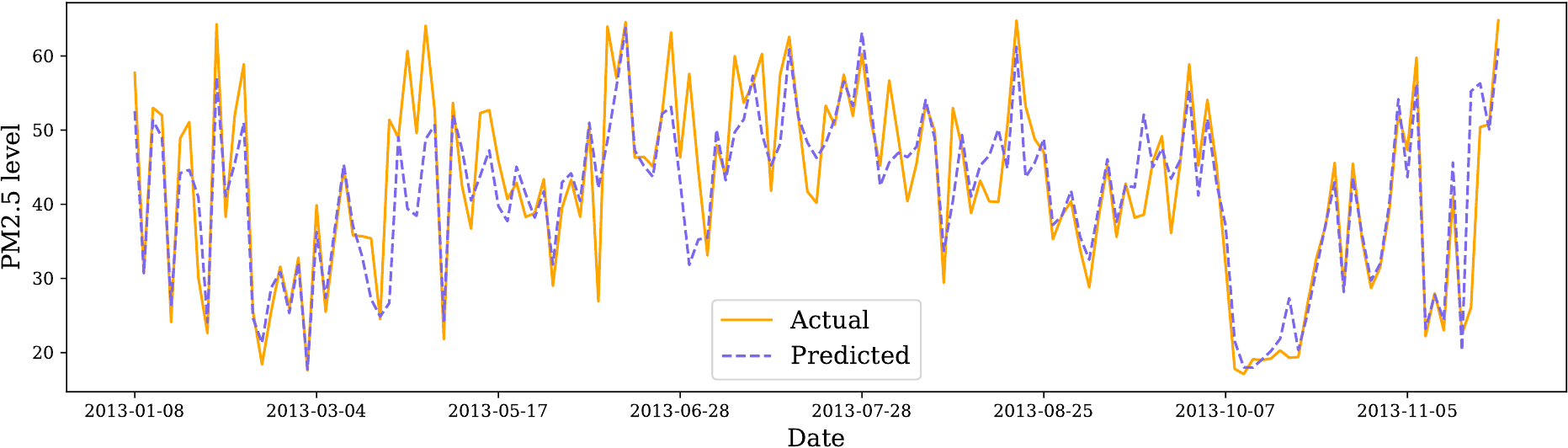}}
		\centerline{(a) Station Number: 1, Location: North-east of Tehran, Period: Jan. 2013 - Jan. 2014}\medskip
	\end{minipage}
	
	\begin{minipage}[b]{1\linewidth} 
		\centering
		\centerline{\includegraphics[width=1\columnwidth]{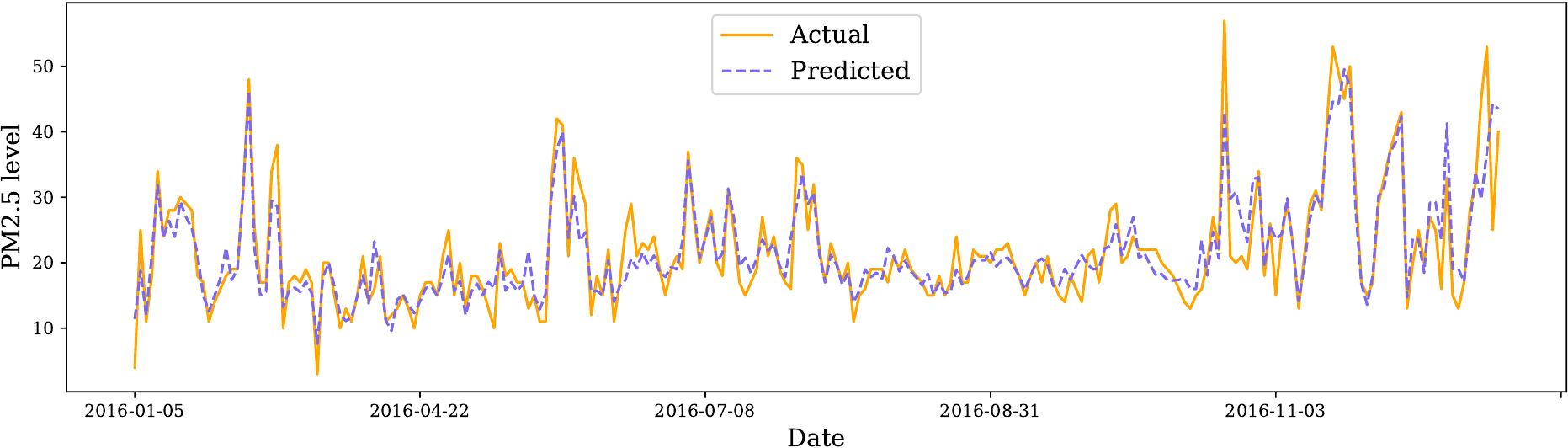}}
		\centerline{(b) Station Number: 15, Location: South of city, Period: Jan. 2016 - Jan. 2017}\medskip 
	\end{minipage}
	
	\begin{minipage}[b]{1\linewidth} 
		\centering
		\centerline{\includegraphics[width=1\columnwidth]{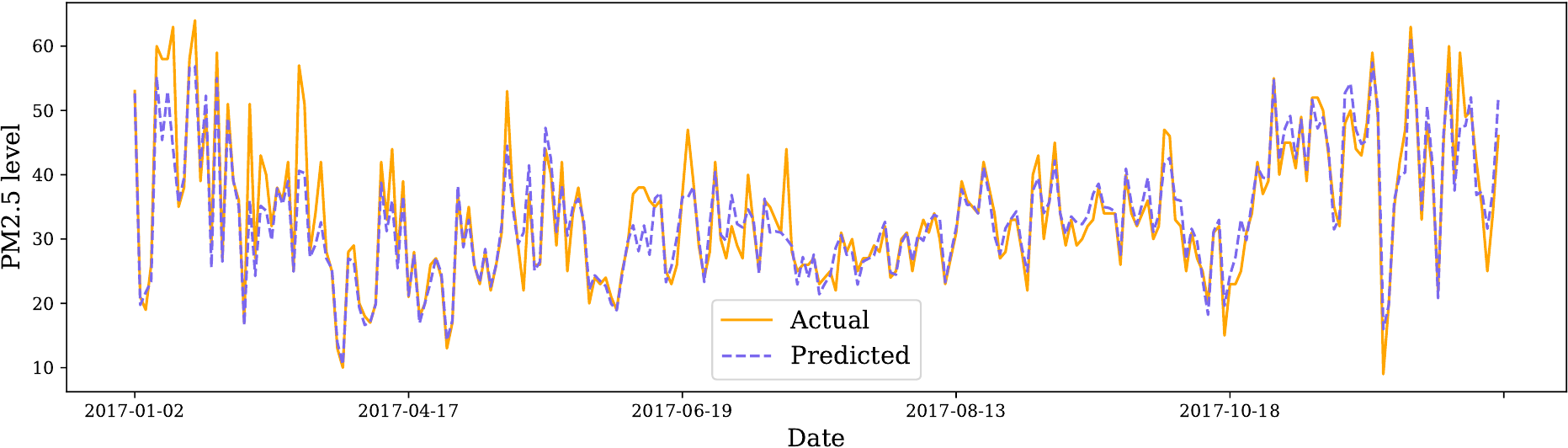}}
		\centerline{(c) Station Number: 22, Location: Center of city, Period: Jan. 2017 - Jan. 2018}\medskip 
	\end{minipage}
	
	\begin{minipage}[b]{1\linewidth} 
		\centering
		\centerline{\includegraphics[width=1\columnwidth]{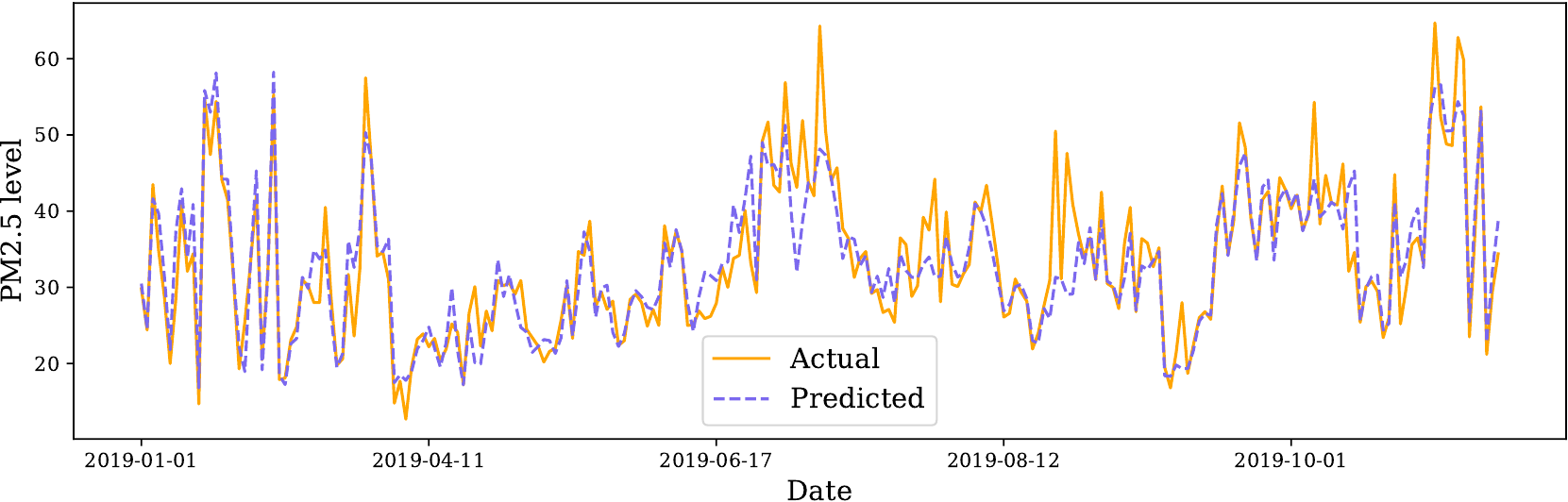}}
		\centerline{(d) Station Number: 7, Location: West of city, Period: Jan. 2019 - Jan. 2020}\medskip 
	\end{minipage}
	\caption{Time series of PM2.5 estimation by the tuned deep ensemble forest at different exemplary stations in versatile locations and in different dates}
	\label{series}
\end{figure}

Fig. \ref{series} visualizes times series of estimations of PM2.5 by the developed deep ensemble forest model at some exemplary air quality monitoring stations for different periods of study dates. At each geographical position, north-east; west; south; center, one station was selected as a sample to validate the accuracy of the tuned model for PM2.5 estimation. For comparison, time series of actual values were also depicted. Time series demonstrate that the deep ensemble forest model can estimate PM2.5 value correctly in most dates of observation. However, for some dates with peak values, the algorithm underestimates the level of PM2.5. This underestimation is also illustrated by regression plots in Fig. \ref{perfomance}. The estimation of PM2.5 by the deep ensemble forest model becomes lower than in-situ measurements for a higher level of concentration. In conclusion, as shown in Fig. \ref{series}, the deep ensemble forest can estimate PM2.5 in accordance with actual measurements in different dates and locations, generally.  

From the achieved results,  it can be concluded that deep learning techniques can exploit useful features through a deep model structure. Nevertheless, the strength of ensemble approaches in dealing with structured data such as PM2.5 estimation from AOD data has been proved in different investigations as mentioned in Section \ref{sec.intro}. Consequently, the deep ensemble forest algorithm by combining useful characteristics of deep learning and decision tree ensemble approaches such as layer-by-layer processing, feature transformation with a deep structure, and learning by the ensemble of ensembles could lead to better results while does not need any backpropagations for training.

\subsection{PM2.5 Concentration Mapping Using Deep Ensemble Forest}\label{map}
After developing the deep ensemble forest model, the tuned model was applied for high resolution (1 $km$) mapping of PM2.5 over the Tehran city. First, the daily PM2.5 map was produced using the input features (AODs from MAIAC, meteorological data, and others). Then, missing PM2.5 values were filled using kriging interpolation technique. 

\begin{figure}[H]
	\centering
	\begin{minipage}[b]{0.4\linewidth} 
		\centering
		\centerline{\includegraphics[width=1\columnwidth]{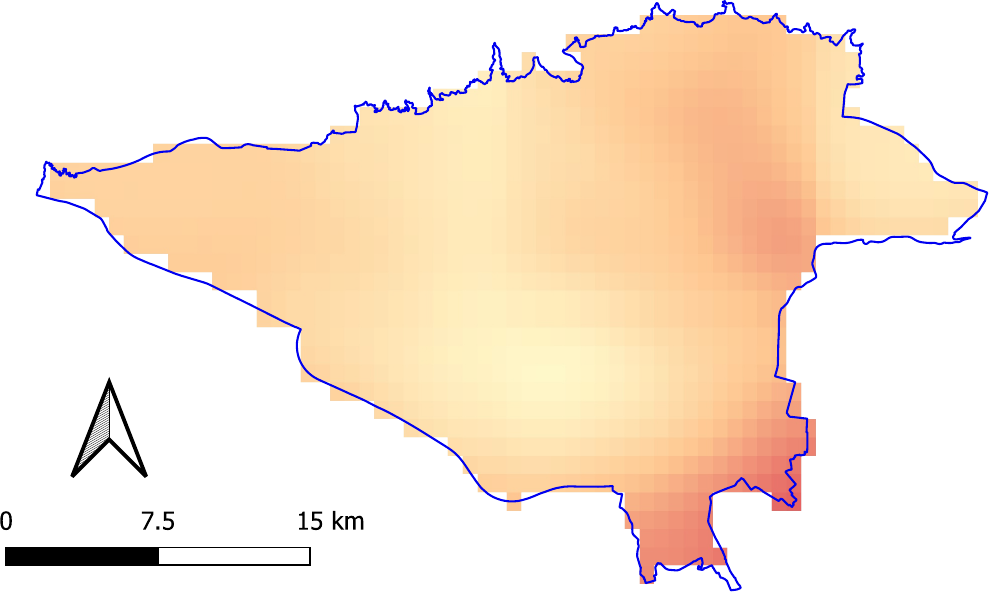}}
		\vspace{0.05cm}
		\centerline{(a) Jan. 1, 2018, AQI: Unhealthy}\medskip	 
	\end{minipage}
	\vspace{0.05cm}
	\hspace{0.5cm}
	\begin{minipage}[b]{0.4\linewidth} 
		\centering
		\centerline{\includegraphics[width=1\columnwidth]{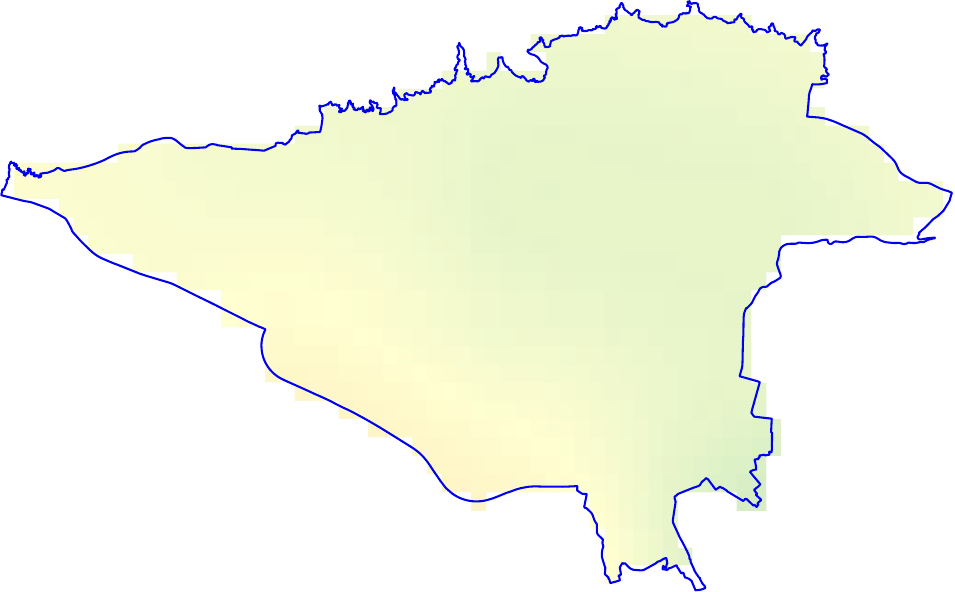}}
		\vspace{0.05cm}
		\centerline{(b) Feb. 18, 2018, AQI: Moderate}\medskip
	\end{minipage}
	\vspace{0.05cm}
	
	\begin{minipage}[b]{0.60\linewidth} 
		\centering
		\centerline{\includegraphics[width=1\columnwidth]{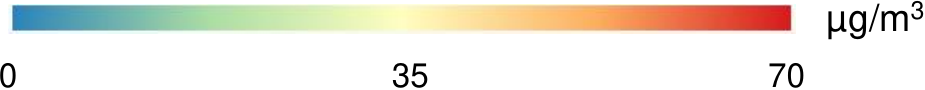}}
		
	\end{minipage}
	\caption{Daily, high resolution PM2.5 map of Tehran at a) Jan. 1, 2018 and b) Feb. 18, 2018.
	}
	\label{daily_map}
\end{figure}

Fig. \ref{daily_map} displays the PM2.5 map of Tehran city in two days with different levels of PM2.5 concentration. One date (Jan. 1, 2018) is labeled as "Unhealthy" and another date (Feb. 18, 2018) is "Moderate" according to the report of AQCC of Tehran. As shown in Fig \ref{daily_map}.a, predicted PM2.5 values by deep ensemble forest in most of the locations have a range between 35 and 70 $\frac{ \mu  g}{ m^{3}}$. Statically, the maximum PM2.5 value predicted for the Jan. 1 is almost 69 $\frac{ \mu  g}{ m^{3}}$. According to Tab. \ref{Tab.pmindex} which lists the air quality index based on the range of PM2.5, the predicted PM2.5 values for Tehran at 1th of Jan. can be lay in "Unhealthy" category since most of the areas have a concentration of more than 55 $\frac{ \mu  g}{ m^{3}}$ and less than 70 $\frac{ \mu  g}{ m^{3}}$ (See map \ref{daily_map}.a). On the 18th of Feb., the achieved map (Fig. \ref{daily_map}.b) shows that most areas have PM2.5 concentration less than 35 $\frac{ \mu  g}{ m^{3}}$. Thus, this date can be labeled as "Moderate" according to categories of air quality index (represented in Tab. \ref{Tab.pmindex}) that is consistent with the report of AQCC. 
In conclusion,  the developed deep ensemble forest can produce high resolution (1 $km$) PM2.5 map with sufficient accuracy that can predict the air quality index over the study area precisely.  

\begin{table}[bt!]
	\centering \footnotesize
	\caption{Categorization of air quality index based on the level of PM2.5}
	\label{Tab.pmindex}
	\begin{tabular}{c c}
		PM2.5        & Air Quality Index \\
		\toprule		
		0 to 12.0    & Good (0 to 50)\\
		12.1 to 35.4 & Moderate (51 to 100)\\ 
		35.5 to 55.4 & Unhealthy for Sensitive Groups (101 to 150)\\
		55.5 to 150.4& Unhealthy (151 to 200)\\
		150.5 to 250.4           & Very Unhealthy (201 to 300)\\
		250.5 to 500.4           & Hazardous (301 to 500)\\
	\end{tabular} 	
\end{table}

\begin{figure}[H]
	\centering
	\begin{minipage}[b]{0.4\linewidth} 
		\centering
		\centerline{\includegraphics[width=1\columnwidth]{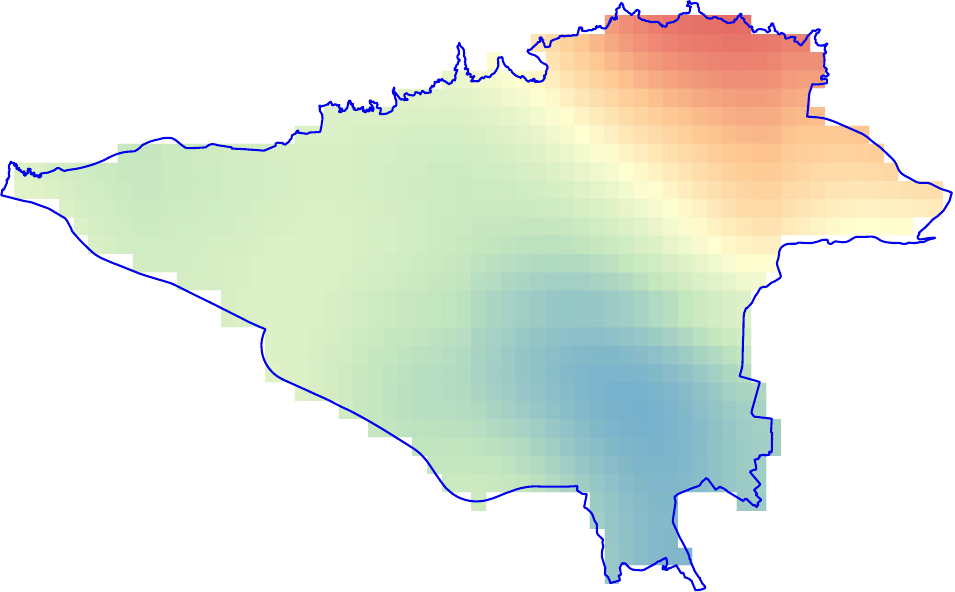}}
		\centerline{(a) 2013}\medskip
	\end{minipage}
	\hspace{0.5cm}
	\vspace{0.05cm}
	\begin{minipage}[b]{0.4\linewidth} 
		\centering
		\centerline{\includegraphics[width=1\columnwidth]{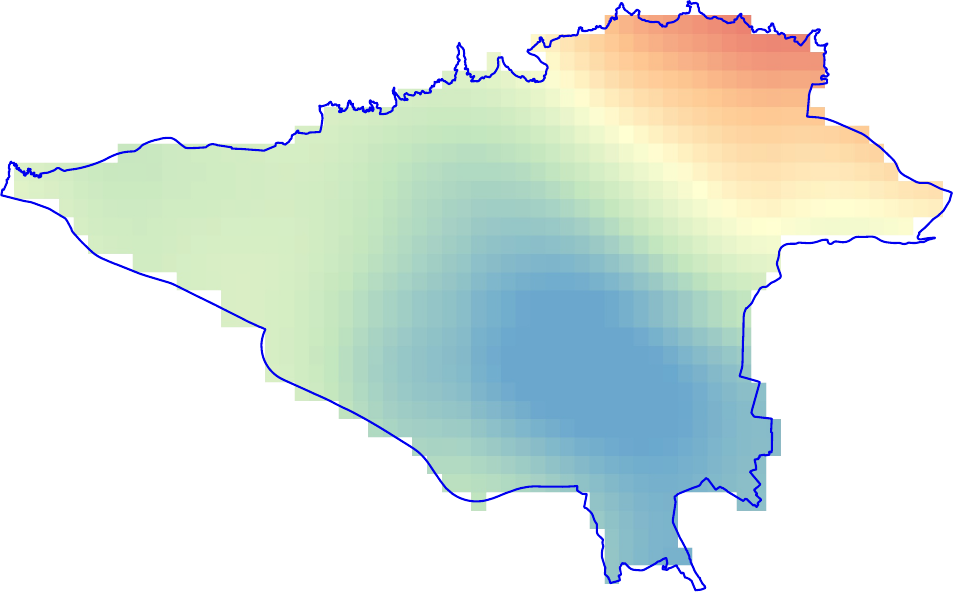}}
		\centerline{(b) 2014}\medskip
	\end{minipage}
	\vspace{0.05cm}
	
	\begin{minipage}[b]{0.4\linewidth} 
		\centering
		\centerline{\includegraphics[width=1\columnwidth]{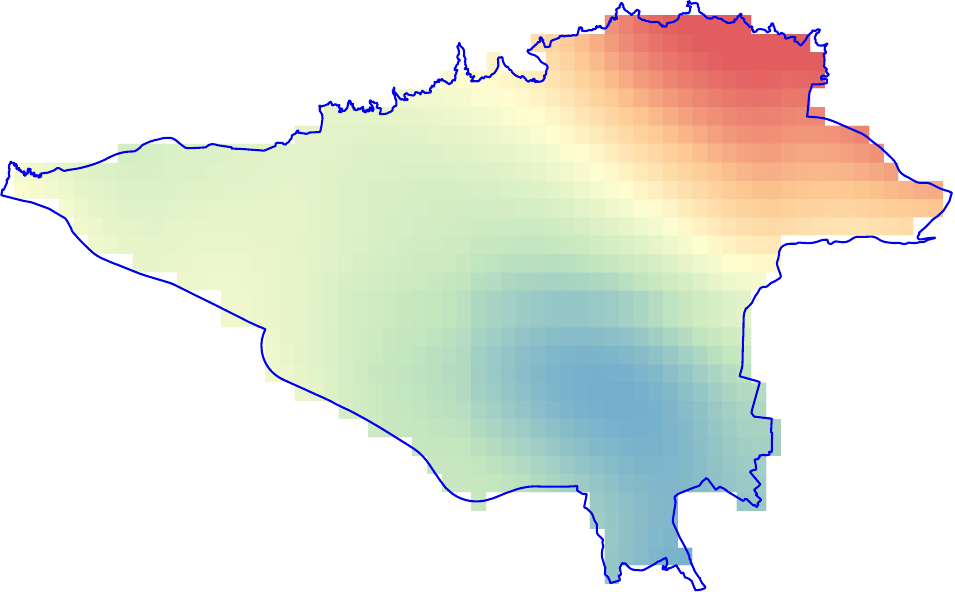}}
		\centerline{(c) 2015}\medskip
	\end{minipage}
	\hspace{0.5cm}
	\vspace{0.05cm}
	\begin{minipage}[b]{0.4\linewidth} 
		\centering
		\centerline{\includegraphics[width=1\columnwidth]{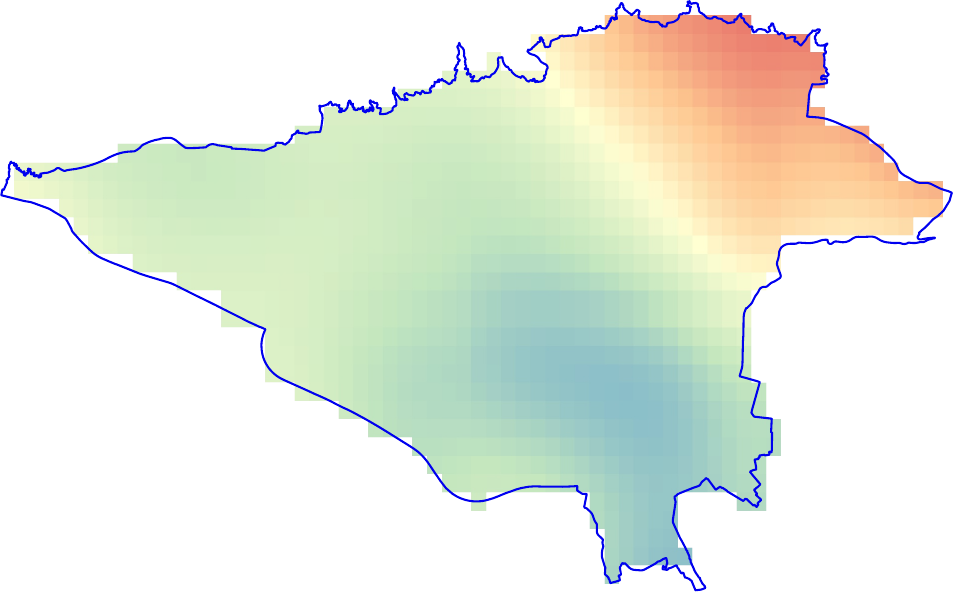}}
		\centerline{(d) 2016}\medskip
	\end{minipage}
	\vspace{0.05cm}
	
	\begin{minipage}[b]{0.4\linewidth} 
		\centering
		\centerline{\includegraphics[width=1\columnwidth]{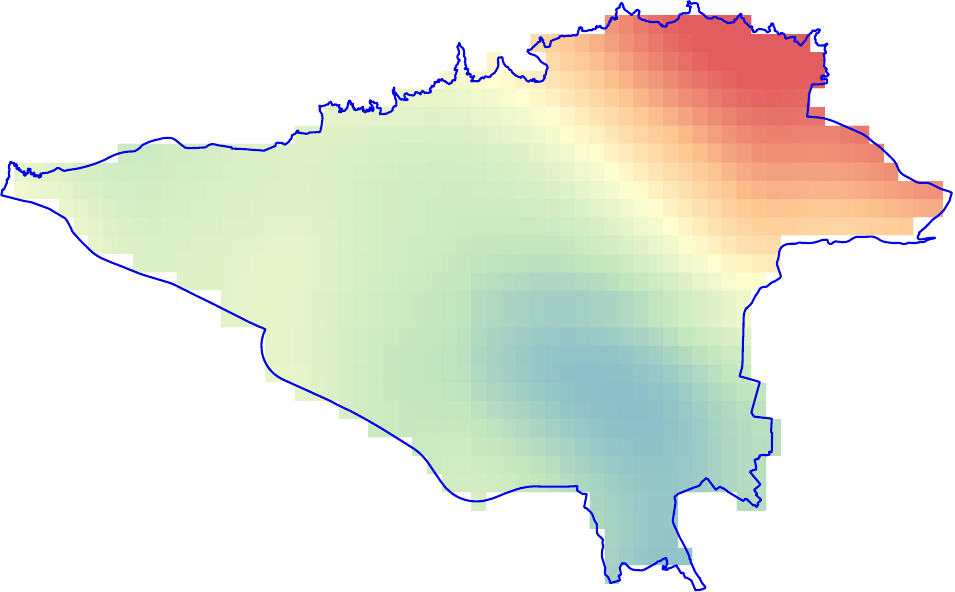}}
		\centerline{(e) 2017}\medskip
	\end{minipage}
	\hspace{0.5cm}
	\vspace{0.05cm}
	\begin{minipage}[b]{0.4\linewidth} 
		\centering
		\centerline{\includegraphics[width=1\columnwidth]{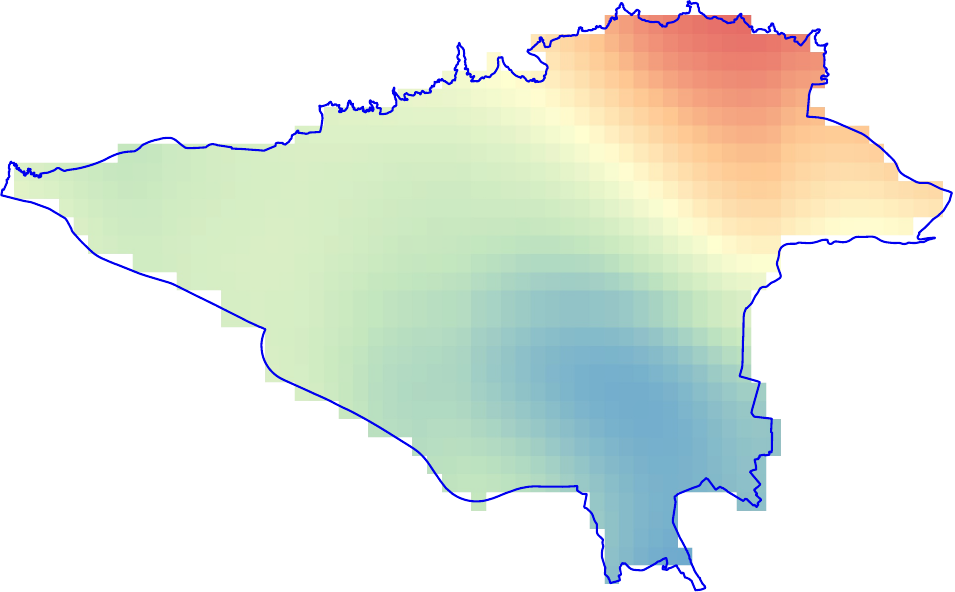}}
		\centerline{(f) 2018}\medskip
	\end{minipage}
	\vspace{0.05cm}
	
	\begin{minipage}[b]{0.4\linewidth} 
		\centering
		\centerline{\includegraphics[width=1\columnwidth]{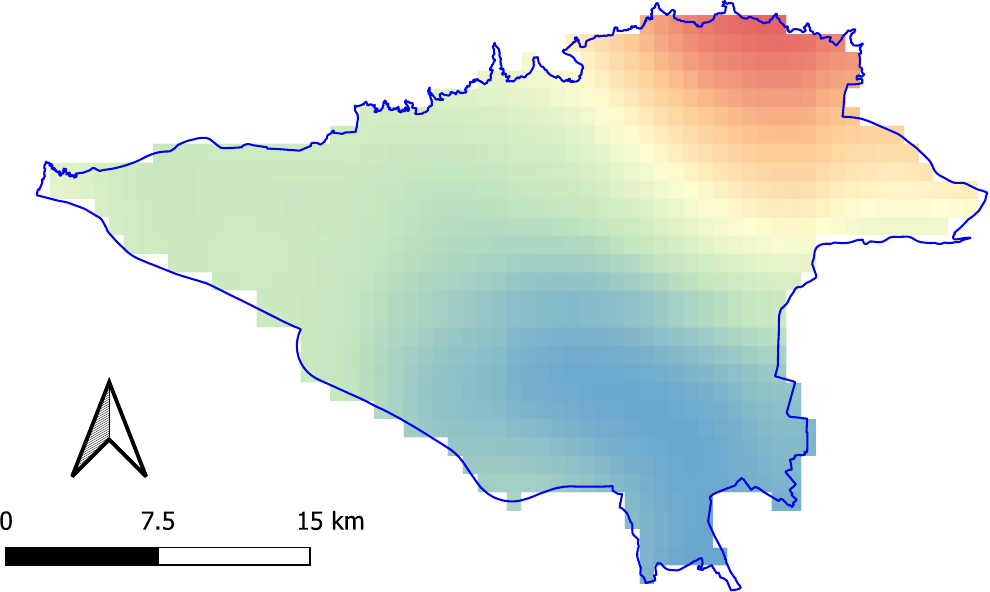}}
		\centerline{(g) 2019}\medskip
	\end{minipage}
	\vspace{0.05cm}

	\begin{minipage}[b]{0.60\linewidth} 
		\centering
		\centerline{\includegraphics[width=1\columnwidth]{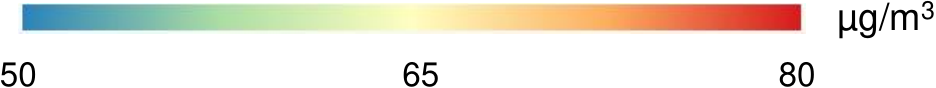}}
		
	\end{minipage}
	\caption{Annual high resolution (1 $km$) maps of PM2.5 over Tehran city 
	}
	\label{map_pm}
\end{figure}

The performance of deep ensemble forest was also verified by the annual estimation of PM2.5 concentration over the study area. In more detail, the yearly average of daily PM2.5 was led to an annual map of PM2.5. Fig. \ref{map_pm} displays annual PM2.5 maps at 1 $km$ realized by the deep ensemble forest model for years between 2013 and 2019.
As shown by the annual PM2.5 maps, the concentration level of pollution alters between around 50 $\frac{ \mu  g}{ m^{3}}$ and 80 $\frac{ \mu  g}{ m^{3}} $.
All generated maps through the deep ensemble forest deployment have almost similar patterns of pollution dispersion. In all maps, the pollution is concentrated in the northeast of Tehran. The results comply with the consequences of \cite{ATASH2007399}. 

\begin{figure}[H]
	\centering
	\begin{minipage}[center]{1\linewidth} 
		\centering
		\centerline{\includegraphics[width=1\columnwidth]{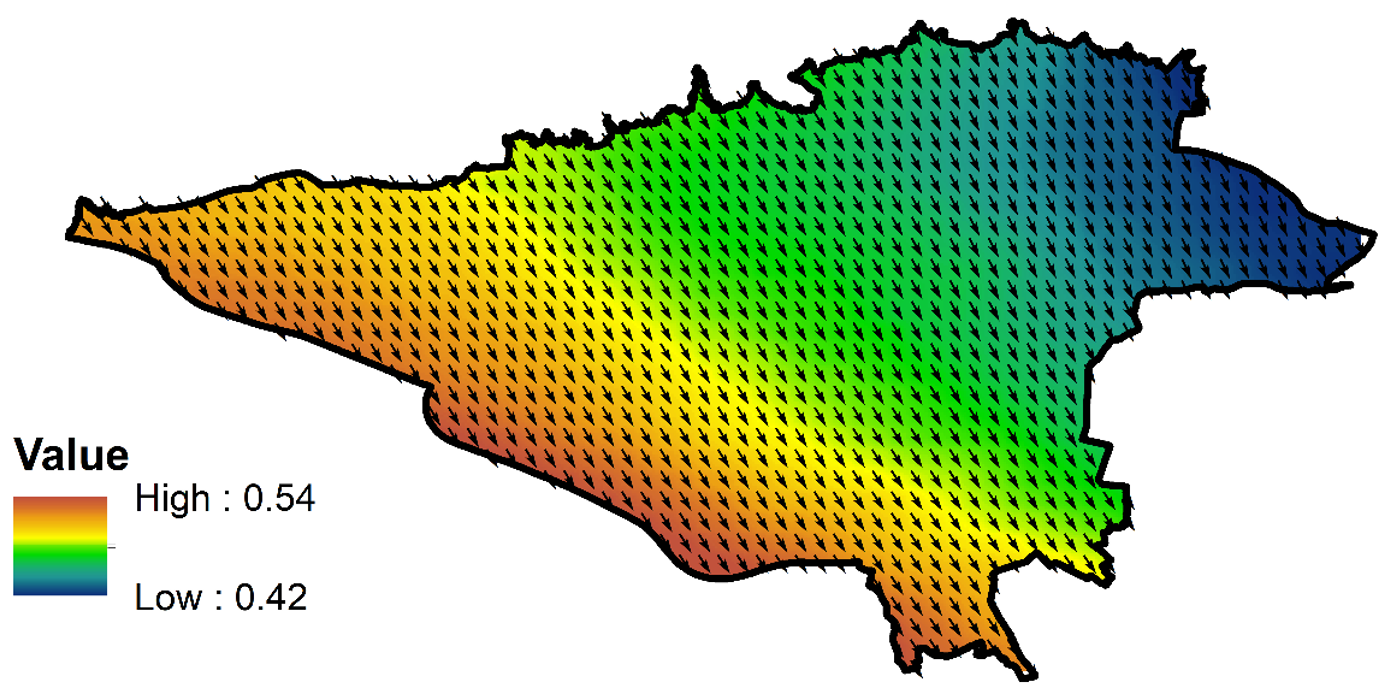}}
	\end{minipage}
	\caption{Annual map of wind directions and speeds ($\frac{m}{s}$) over the study area in 2013
	}
	\label{map_wind}
\end{figure}

\begin{figure}[H]
	\centering
	\begin{minipage}[b]{0.7\linewidth} 
		\centering
		\centerline{\includegraphics[width=1\columnwidth]{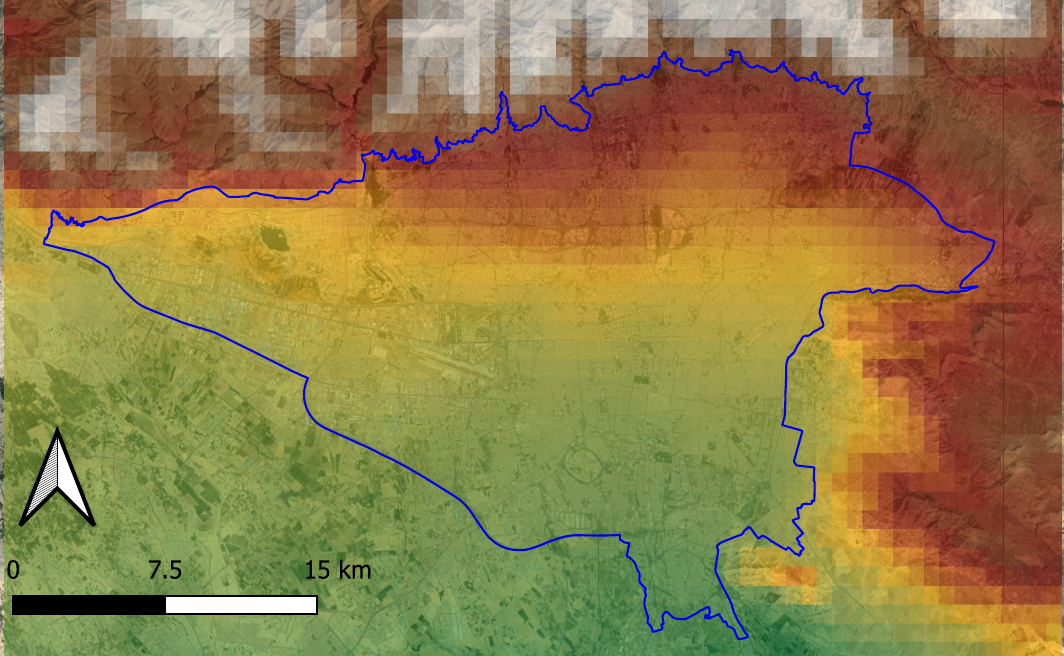}}
	\end{minipage}
	\vspace{0.3cm}
	
	\begin{minipage}[b]{0.5\linewidth} 
		\centering
		\centerline{\includegraphics[width=1\columnwidth]{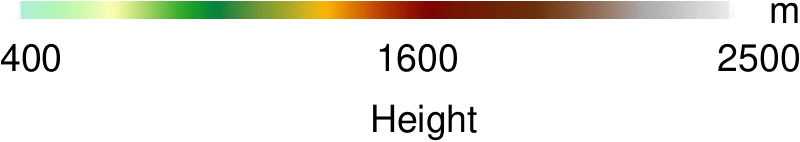}}
	\end{minipage}
	\caption{Topography map of Tehran and surroundings.
	}
	\label{map_topo}
\end{figure}

For a better interpretation of generated maps, the corresponding annual wind maps, including directions and speeds of wind vectors as well as the topography map of the study area are depicted in Fig. \ref{map_wind} and \ref{map_topo}, respectively.

As shown in wind map, all winds have almost the same direction from the northwest to the southeast. However, the wind speed decreases from the west to the east, where it reaches the minimum speed, 0.42 $m/s$. On the other hand, based on the topography representation, the east and the northeast are surrounded by the highest mountains. As a result, increasing the height of mountains as well as decreasing the wind speed in the east and the northeast of the city cause trapping pollutants and subsequently concentrating PM2.5 in this area more than other parts of the city. 

\section{Conclusion}\label{conclude}

This paper investigates the potential of deep ensemble forest for estimating PM2.5 concentration from satellite AOD data (MAIAC MODIS) and other parameters such as meteorological observations. The goal is to produce a high resolution (1 $km$) PM2.5 map in the study area, Tehran. For this, deep ensemble forest is developed to explore a regression function that will be used for PM2.5 estimation. The results of implemented experiments in this paper demonstrated the ability of deep ensemble forest in PM2.5 estimation. The algorithm was also compared with other machine learning techniques such as random forest, deep neural networks, etc. The comparison revealed that the developed deep ensemble forest could outperform other methods such as deep neural networks.

\section*{Acknowledgments}
The author wishes to express his gratitude to everyone who contributed to this study, particularly Tehran's AQCC for the ground PM2.5 measurements, NASA for the MAIAC data, and ECMWF for the meteorological data.

\section*{Declarations}

\subsection*{Ethical Approval}

All authors have read, understood, and have complied as applicable with the statement on "Ethical responsibilities of Authors" as found in the Instructions for Authors and are aware that with minor exceptions, no changes can be made to authorship once the paper is submitted.
 
\subsection*{Conflict of Interest} 
The author declares that he has no conflict of interest.
 

\subsection*{Funding}
This research received no specific grant from any funding agency in the public, commercial, or not-for-profit sectors.

\subsection*{Availability of data and materials }
The data that support the findings of this study are available on request from the corresponding author.

\bibliography{sn-bibliography}

\begin{thebibliography}{}
\providecommand{\doi}[1]{\url{https://doi.org/#1}}
\bibcommenthead

\bibitem [\protect \citeauthoryear {%
Ahmad%
\ \protect \BOthers {.}}{%
Ahmad%
\ \protect \BOthers {.}}{%
{\protect \APACyear {2019}}%
}]{%
AHMAD2019117050}
\APACinsertmetastar {%
AHMAD2019117050}%
\begin{APACrefauthors}%
Ahmad, M.%
, Alam, K.%
, Tariq, S.%
, Anwar, S.%
, Nasir, J.%
\BCBL {} Mansha, M.%
\end{APACrefauthors}%
\unskip\
\newblock
\APACrefYearMonthDay{2019}{}{}.
\newblock
{\BBOQ}\APACrefatitle {Estimating fine particulate concentration using a
  combined approach of linear regression and artificial neural network}
  {Estimating fine particulate concentration using a combined approach of
  linear regression and artificial neural network}.{\BBCQ}
\newblock
\APACjournalVolNumPages{Atmospheric Environment}{219}{}{117050}.
\newblock

\newblock

\PrintBackRefs{\CurrentBib}

\bibitem [\protect \citeauthoryear {%
Anguita%
, Ghelardoni%
, Ghio%
, Oneto%
\BCBL {}\ \BBA {} Ridella%
}{%
Anguita%
\ \protect \BOthers {.}}{%
{\protect \APACyear {2012}}%
}]{%
anguita2012k}
\APACinsertmetastar {%
anguita2012k}%
\begin{APACrefauthors}%
Anguita, D.%
, Ghelardoni, L.%
, Ghio, A.%
, Oneto, L.%
\BCBL {} Ridella, S.%
\end{APACrefauthors}%
\unskip\
\newblock
\APACrefYearMonthDay{2012}{}{}.
\newblock
{\BBOQ}\APACrefatitle {The {‘K’}in {K-fold} cross validation} {The
  {‘K’}in {K-fold} cross validation}.{\BBCQ}
\newblock
 \APACrefbtitle {20th European Symposium on Artificial Neural Networks,
  Computational Intelligence and Machine Learning (ESANN)} {20th european
  symposium on artificial neural networks, computational intelligence and
  machine learning (esann)}\ (\BPGS\ 441--446).
\PrintBackRefs{\CurrentBib}

\bibitem [\protect \citeauthoryear {%
Arciszewska%
\ \BBA {} McClatchey%
}{%
Arciszewska%
\ \BBA {} McClatchey%
}{%
{\protect \APACyear {2001}}%
}]{%
arciszewska2001importance}
\APACinsertmetastar {%
arciszewska2001importance}%
\begin{APACrefauthors}%
Arciszewska, C.%
\BCBT {}\ \BBA {} McClatchey, J.%
\end{APACrefauthors}%
\unskip\
\newblock
\APACrefYearMonthDay{2001}{}{}.
\newblock
{\BBOQ}\APACrefatitle {The importance of meteorological data for modelling air
  pollution using {ADMS-Urban}} {The importance of meteorological data for
  modelling air pollution using {ADMS-Urban}}.{\BBCQ}
\newblock
\APACjournalVolNumPages{Meteorological Applications: A journal of forecasting,
  practical applications, training techniques and modelling}{8}{3}{345--350}.
\newblock

\newblock

\PrintBackRefs{\CurrentBib}

\bibitem [\protect \citeauthoryear {%
Atash%
}{%
Atash%
}{%
{\protect \APACyear {2007}}%
}]{%
ATASH2007399}
\APACinsertmetastar {%
ATASH2007399}%
\begin{APACrefauthors}%
Atash, F.%
\end{APACrefauthors}%
\unskip\
\newblock
\APACrefYearMonthDay{2007}{}{}.
\newblock
{\BBOQ}\APACrefatitle {The deterioration of urban environments in developing
  countries: Mitigating the air pollution crisis in {Tehran}, {Iran}} {The
  deterioration of urban environments in developing countries: Mitigating the
  air pollution crisis in {Tehran}, {Iran}}.{\BBCQ}
\newblock
\APACjournalVolNumPages{Cities}{24}{6}{399-409}.
\newblock

\newblock

\PrintBackRefs{\CurrentBib}

\bibitem [\protect \citeauthoryear {%
Bagheri%
}{%
Bagheri%
}{%
{\protect \APACyear {2022}}%
}]{%
bagheri2019ASRJ}
\APACinsertmetastar {%
bagheri2019ASRJ}%
\begin{APACrefauthors}%
Bagheri, H.%
\end{APACrefauthors}%
\unskip\
\newblock
\APACrefYearMonthDay{2022}{}{}.
\newblock
{\BBOQ}\APACrefatitle {A Machine Learning-based Framework for High Resolution
  Mapping of {PM2.5} in {Tehran, Iran}, Using {MAIAC AOD} Data} {A machine
  learning-based framework for high resolution mapping of {PM2.5} in {Tehran,
  Iran}, using {MAIAC AOD} data}.{\BBCQ}
\newblock
\APACjournalVolNumPages{Advances in space Research}{}{}{}.
\newblock
\APACrefnote{In press.}
\newblock

\newblock

\PrintBackRefs{\CurrentBib}

\bibitem [\protect \citeauthoryear {%
Bagheri%
, Sadeghian%
\BCBL {}\ \BBA {} Sadjadi%
}{%
Bagheri%
\ \protect \BOthers {.}}{%
{\protect \APACyear {2014}}%
}]{%
bagheri2014}
\APACinsertmetastar {%
bagheri2014}%
\begin{APACrefauthors}%
Bagheri, H.%
, Sadeghian, S.%
\BCBL {} Sadjadi, S.Y.%
\end{APACrefauthors}%
\unskip\
\newblock
\APACrefYearMonthDay{2014}{}{}.
\newblock
{\BBOQ}\APACrefatitle {The Assessment of using an Intelligent Algorithm for the
  Interpolation of Elevation in the {DTM} Generation} {The assessment of using
  an intelligent algorithm for the interpolation of elevation in the {DTM}
  generation}.{\BBCQ}
\newblock
\APACjournalVolNumPages{Photogrammetrie - Fernerkundung -
  Geoinformation}{2014}{3}{197-208}.
\newblock

\newblock

\PrintBackRefs{\CurrentBib}

\bibitem [\protect \citeauthoryear {%
Buitinck%
\ \protect \BOthers {.}}{%
Buitinck%
\ \protect \BOthers {.}}{%
{\protect \APACyear {2013}}%
}]{%
sklearn_api}
\APACinsertmetastar {%
sklearn_api}%
\begin{APACrefauthors}%
Buitinck, L.%
, Louppe, G.%
, Blondel, M.%
, Pedregosa, F.%
, Mueller, A.%
, Grisel, O.%
\BDBL {}Varoquaux, G.%
\end{APACrefauthors}%
\unskip\
\newblock
\APACrefYearMonthDay{2013}{}{}.
\newblock
{\BBOQ}\APACrefatitle {{API} design for machine learning software: experiences
  from the scikit-learn project} {{API} design for machine learning software:
  experiences from the scikit-learn project}.{\BBCQ}
\newblock
 \APACrefbtitle {ECML PKDD Workshop: Languages for Data Mining and Machine
  Learning} {Ecml pkdd workshop: Languages for data mining and machine
  learning}\ (\BPGS\ 108--122).
\PrintBackRefs{\CurrentBib}

\bibitem [\protect \citeauthoryear {%
B.~Chen%
\ \protect \BOthers {.}}{%
B.~Chen%
\ \protect \BOthers {.}}{%
{\protect \APACyear {2021}}%
}]{%
CHEN2021144724}
\APACinsertmetastar {%
CHEN2021144724}%
\begin{APACrefauthors}%
Chen, B.%
, You, S.%
, Ye, Y.%
, Fu, Y.%
, Ye, Z.%
, Deng, J.%
\BDBL {}Hong, Y.%
\end{APACrefauthors}%
\unskip\
\newblock
\APACrefYearMonthDay{2021}{}{}.
\newblock
{\BBOQ}\APACrefatitle {An interpretable self-adaptive deep neural network for
  estimating daily spatially-continuous {PM2.5} concentrations across {China}}
  {An interpretable self-adaptive deep neural network for estimating daily
  spatially-continuous {PM2.5} concentrations across {China}}.{\BBCQ}
\newblock
\APACjournalVolNumPages{Science of The Total Environment}{768}{}{144724}.
\newblock

\newblock

\PrintBackRefs{\CurrentBib}

\bibitem [\protect \citeauthoryear {%
W.~Chen%
\ \protect \BOthers {.}}{%
W.~Chen%
\ \protect \BOthers {.}}{%
{\protect \APACyear {2020}}%
}]{%
chen2020estimating}
\APACinsertmetastar {%
chen2020estimating}%
\begin{APACrefauthors}%
Chen, W.%
, Ran, H.%
, Cao, X.%
, Wang, J.%
, Teng, D.%
, Chen, J.%
\BCBL {} Zheng, X.%
\end{APACrefauthors}%
\unskip\
\newblock
\APACrefYearMonthDay{2020}{}{}.
\newblock
{\BBOQ}\APACrefatitle {Estimating {PM2.5} with high-resolution 1-km {AOD} data
  and an improved machine learning model over {Shenzhen, China}} {Estimating
  {PM2.5} with high-resolution 1-km {AOD} data and an improved machine learning
  model over {Shenzhen, China}}.{\BBCQ}
\newblock
\APACjournalVolNumPages{Science of The Total Environment}{746}{}{141093}.
\newblock

\newblock

\PrintBackRefs{\CurrentBib}

\bibitem [\protect \citeauthoryear {%
Di%
\ \protect \BOthers {.}}{%
Di%
\ \protect \BOthers {.}}{%
{\protect \APACyear {2016}}%
}]{%
di2016assessing}
\APACinsertmetastar {%
di2016assessing}%
\begin{APACrefauthors}%
Di, Q.%
, Kloog, I.%
, Koutrakis, P.%
, Lyapustin, A.%
, Wang, Y.%
\BCBL {} Schwartz, J.%
\end{APACrefauthors}%
\unskip\
\newblock
\APACrefYearMonthDay{2016}{}{}.
\newblock
{\BBOQ}\APACrefatitle {Assessing {PM2.5} exposures with high spatiotemporal
  resolution across the continental {United States}} {Assessing {PM2.5}
  exposures with high spatiotemporal resolution across the continental {United
  States}}.{\BBCQ}
\newblock
\APACjournalVolNumPages{Environmental science \&
  technology}{50}{9}{4712--4721}.
\newblock

\newblock

\PrintBackRefs{\CurrentBib}

\bibitem [\protect \citeauthoryear {%
ECMWF%
}{%
ECMWF%
}{%
{\protect \APACyear {2021}}%
}]{%
ER5}
\APACinsertmetastar {%
ER5}%
\begin{APACrefauthors}%
ECMWF%
\end{APACrefauthors}%
\unskip\
\newblock
\APACrefYearMonthDay{2021}{}{}.
\newblock
\APACrefbtitle {{ERA5}.} {{ERA5}.}
\newblock
\APAChowpublished {\url{https://confluence.ecmwf.int/display/CKB/ERA5}}.
\newblock
\APACrefnote{[Accessed 02.21]}
\PrintBackRefs{\CurrentBib}

\bibitem [\protect \citeauthoryear {%
Geurts%
, Ernst%
\BCBL {}\ \BBA {} Wehenkel%
}{%
Geurts%
\ \protect \BOthers {.}}{%
{\protect \APACyear {2006}}%
}]{%
geurts2006extremely}
\APACinsertmetastar {%
geurts2006extremely}%
\begin{APACrefauthors}%
Geurts, P.%
, Ernst, D.%
\BCBL {} Wehenkel, L.%
\end{APACrefauthors}%
\unskip\
\newblock
\APACrefYearMonthDay{2006}{}{}.
\newblock
{\BBOQ}\APACrefatitle {Extremely randomized trees} {Extremely randomized
  trees}.{\BBCQ}
\newblock
\APACjournalVolNumPages{Machine learning}{63}{1}{3--42}.
\newblock

\newblock

\PrintBackRefs{\CurrentBib}

\bibitem [\protect \citeauthoryear {%
Ghotbi%
, Sotoudeheian%
\BCBL {}\ \BBA {} Arhami%
}{%
Ghotbi%
\ \protect \BOthers {.}}{%
{\protect \APACyear {2016}}%
}]{%
GHOTBI2016333}
\APACinsertmetastar {%
GHOTBI2016333}%
\begin{APACrefauthors}%
Ghotbi, S.%
, Sotoudeheian, S.%
\BCBL {} Arhami, M.%
\end{APACrefauthors}%
\unskip\
\newblock
\APACrefYearMonthDay{2016}{}{}.
\newblock
{\BBOQ}\APACrefatitle {Estimating urban ground-level {PM10} using {MODIS} 3km
  {AOD} product and meteorological parameters from {WRF} model} {Estimating
  urban ground-level {PM10} using {MODIS} 3km {AOD} product and meteorological
  parameters from {WRF} model}.{\BBCQ}
\newblock
\APACjournalVolNumPages{Atmospheric Environment}{141}{}{333-346}.
\newblock

\newblock

\PrintBackRefs{\CurrentBib}

\bibitem [\protect \citeauthoryear {%
Goodfellow%
, Bengio%
, Courville%
\BCBL {}\ \BBA {} Bengio%
}{%
Goodfellow%
\ \protect \BOthers {.}}{%
{\protect \APACyear {2016}}%
}]{%
goodfellow2016deep}
\APACinsertmetastar {%
goodfellow2016deep}%
\begin{APACrefauthors}%
Goodfellow, I.%
, Bengio, Y.%
, Courville, A.%
\BCBL {} Bengio, Y.%
\end{APACrefauthors}%
\unskip\
\newblock
\APACrefYear{2016}.
\newblock
\APACrefbtitle {Deep learning} {Deep learning}\ (\BVOL~1)\ (\BNUM~2).
\newblock
\APACaddressPublisher{}{MIT press Cambridge}.
\PrintBackRefs{\CurrentBib}

\bibitem [\protect \citeauthoryear {%
Gupta%
\ \BBA {} Christopher%
}{%
Gupta%
\ \BBA {} Christopher%
}{%
{\protect \APACyear {2009}}%
{\protect \APACexlab {{\protect \BCnt {1}}}}}]{%
https://doi.org/10.1029/2008JD011497}
\APACinsertmetastar {%
https://doi.org/10.1029/2008JD011497}%
\begin{APACrefauthors}%
Gupta, P.%
\BCBT {}\ \BBA {} Christopher, S.A.%
\end{APACrefauthors}%
\unskip\
\newblock
\APACrefYearMonthDay{2009{\protect \BCnt {1}}}{}{}.
\newblock
{\BBOQ}\APACrefatitle {Particulate matter air quality assessment using
  integrated surface, satellite, and meteorological products: 2. {A} neural
  network approach} {Particulate matter air quality assessment using integrated
  surface, satellite, and meteorological products: 2. {A} neural network
  approach}.{\BBCQ}
\newblock
\APACjournalVolNumPages{Journal of Geophysical Research:
  Atmospheres}{114}{D20}{}.
\newblock

\newblock

\PrintBackRefs{\CurrentBib}

\bibitem [\protect \citeauthoryear {%
Gupta%
\ \BBA {} Christopher%
}{%
Gupta%
\ \BBA {} Christopher%
}{%
{\protect \APACyear {2009}}%
{\protect \APACexlab {{\protect \BCnt {2}}}}}]{%
gupta_regression}
\APACinsertmetastar {%
gupta_regression}%
\begin{APACrefauthors}%
Gupta, P.%
\BCBT {}\ \BBA {} Christopher, S.A.%
\end{APACrefauthors}%
\unskip\
\newblock
\APACrefYearMonthDay{2009{\protect \BCnt {2}}}{}{}.
\newblock
{\BBOQ}\APACrefatitle {Particulate matter air quality assessment using
  integrated surface, satellite, and meteorological products: Multiple
  regression approach} {Particulate matter air quality assessment using
  integrated surface, satellite, and meteorological products: Multiple
  regression approach}.{\BBCQ}
\newblock
\APACjournalVolNumPages{Journal of Geophysical Research:
  Atmospheres}{114}{D14}{}.
\newblock

\newblock

\PrintBackRefs{\CurrentBib}

\bibitem [\protect \citeauthoryear {%
Gupta%
\ \BBA {} Christopher%
}{%
Gupta%
\ \BBA {} Christopher%
}{%
{\protect \APACyear {2009}}%
{\protect \APACexlab {{\protect \BCnt {3}}}}}]{%
https://doi.org/10.1029/2008JD011496}
\APACinsertmetastar {%
https://doi.org/10.1029/2008JD011496}%
\begin{APACrefauthors}%
Gupta, P.%
\BCBT {}\ \BBA {} Christopher, S.A.%
\end{APACrefauthors}%
\unskip\
\newblock
\APACrefYearMonthDay{2009{\protect \BCnt {3}}}{}{}.
\newblock
{\BBOQ}\APACrefatitle {Particulate matter air quality assessment using
  integrated surface, satellite, and meteorological products: Multiple
  regression approach} {Particulate matter air quality assessment using
  integrated surface, satellite, and meteorological products: Multiple
  regression approach}.{\BBCQ}
\newblock
\APACjournalVolNumPages{Journal of Geophysical Research:
  Atmospheres}{114}{D14}{}.
\newblock

\newblock

\PrintBackRefs{\CurrentBib}

\bibitem [\protect \citeauthoryear {%
Hersbach%
\ \protect \BOthers {.}}{%
Hersbach%
\ \protect \BOthers {.}}{%
{\protect \APACyear {2020}}%
}]{%
https://doi.org/10.1002/qj.3803}
\APACinsertmetastar {%
https://doi.org/10.1002/qj.3803}%
\begin{APACrefauthors}%
Hersbach, H.%
, Bell, B.%
, Berrisford, P.%
, Hirahara, S.%
, Horányi, A.%
, Muñoz-Sabater, J.%
\BDBL {}Thépaut, J\BHBI N.%
\end{APACrefauthors}%
\unskip\
\newblock
\APACrefYearMonthDay{2020}{}{}.
\newblock
{\BBOQ}\APACrefatitle {The {ERA5} global reanalysis} {The {ERA5} global
  reanalysis}.{\BBCQ}
\newblock
\APACjournalVolNumPages{Quarterly Journal of the Royal Meteorological
  Society}{146}{730}{1999-2049}.
\newblock

\newblock

\PrintBackRefs{\CurrentBib}

\bibitem [\protect \citeauthoryear {%
Hinton%
, Osindero%
\BCBL {}\ \BBA {} Teh%
}{%
Hinton%
\ \protect \BOthers {.}}{%
{\protect \APACyear {2006}}%
}]{%
hinton2006fast}
\APACinsertmetastar {%
hinton2006fast}%
\begin{APACrefauthors}%
Hinton, G.E.%
, Osindero, S.%
\BCBL {} Teh, Y\BHBI W.%
\end{APACrefauthors}%
\unskip\
\newblock
\APACrefYearMonthDay{2006}{}{}.
\newblock
{\BBOQ}\APACrefatitle {A fast learning algorithm for deep belief nets} {A fast
  learning algorithm for deep belief nets}.{\BBCQ}
\newblock
\APACjournalVolNumPages{Neural computation}{18}{7}{1527--1554}.
\newblock

\newblock

\PrintBackRefs{\CurrentBib}

\bibitem [\protect \citeauthoryear {%
Hsu%
\ \protect \BOthers {.}}{%
Hsu%
\ \protect \BOthers {.}}{%
{\protect \APACyear {2013}}%
}]{%
sayeretal}
\APACinsertmetastar {%
sayeretal}%
\begin{APACrefauthors}%
Hsu, N.C.%
, Jeong, M\BHBI J.%
, Bettenhausen, C.%
, Sayer, A.M.%
, Hansell, R.%
, Seftor, C.S.%
\BDBL {}Tsay, S\BHBI C.%
\end{APACrefauthors}%
\unskip\
\newblock
\APACrefYearMonthDay{2013}{}{}.
\newblock
{\BBOQ}\APACrefatitle {Enhanced Deep Blue aerosol retrieval algorithm: The
  second generation} {Enhanced deep blue aerosol retrieval algorithm: The
  second generation}.{\BBCQ}
\newblock
\APACjournalVolNumPages{Journal of Geophysical Research:
  Atmospheres}{118}{16}{9296-9315}.
\newblock

\newblock

\PrintBackRefs{\CurrentBib}

\bibitem [\protect \citeauthoryear {%
Hu%
\ \protect \BOthers {.}}{%
Hu%
\ \protect \BOthers {.}}{%
{\protect \APACyear {2014}}%
}]{%
hu2014estimating}
\APACinsertmetastar {%
hu2014estimating}%
\begin{APACrefauthors}%
Hu, X.%
, Waller, L.A.%
, Lyapustin, A.%
, Wang, Y.%
, Al-Hamdan, M.Z.%
, Crosson, W.L.%
\BDBL {}others%
\end{APACrefauthors}%
\unskip\
\newblock
\APACrefYearMonthDay{2014}{}{}.
\newblock
{\BBOQ}\APACrefatitle {Estimating ground-level {PM2.5} concentrations in the
  {Southeastern United States} using {MAIAC AOD} retrievals and a two-stage
  model} {Estimating ground-level {PM2.5} concentrations in the {Southeastern
  United States} using {MAIAC AOD} retrievals and a two-stage model}.{\BBCQ}
\newblock
\APACjournalVolNumPages{Remote Sensing of Environment}{140}{}{220--232}.
\newblock

\newblock

\PrintBackRefs{\CurrentBib}

\bibitem [\protect \citeauthoryear {%
James%
, Witten%
, Hastie%
\BCBL {}\ \BBA {} Tibshirani%
}{%
James%
\ \protect \BOthers {.}}{%
{\protect \APACyear {2013}}%
}]{%
james2013introduction}
\APACinsertmetastar {%
james2013introduction}%
\begin{APACrefauthors}%
James, G.%
, Witten, D.%
, Hastie, T.%
\BCBL {} Tibshirani, R.%
\end{APACrefauthors}%
\unskip\
\newblock
\APACrefYear{2013}.
\newblock
\APACrefbtitle {An introduction to statistical learning} {An introduction to
  statistical learning}\ (\BVOL~112).
\newblock
\APACaddressPublisher{}{Springer}.
\PrintBackRefs{\CurrentBib}

\bibitem [\protect \citeauthoryear {%
Klemm%
, Mason~Jr%
, Heilig%
, Neas%
\BCBL {}\ \BBA {} Dockery%
}{%
Klemm%
\ \protect \BOthers {.}}{%
{\protect \APACyear {2000}}%
}]{%
klemm2000daily}
\APACinsertmetastar {%
klemm2000daily}%
\begin{APACrefauthors}%
Klemm, R.J.%
, Mason~Jr, R.M.%
, Heilig, C.M.%
, Neas, L.M.%
\BCBL {} Dockery, D.W.%
\end{APACrefauthors}%
\unskip\
\newblock
\APACrefYearMonthDay{2000}{}{}.
\newblock
{\BBOQ}\APACrefatitle {Is daily mortality associated specifically with fine
  particles? {Data} reconstruction and replication of analyses} {Is daily
  mortality associated specifically with fine particles? {Data} reconstruction
  and replication of analyses}.{\BBCQ}
\newblock
\APACjournalVolNumPages{Journal of the Air \& Waste Management
  Association}{50}{7}{1215--1222}.
\newblock

\newblock

\PrintBackRefs{\CurrentBib}

\bibitem [\protect \citeauthoryear {%
Lee%
, Liu%
, Coull%
, Schwartz%
\BCBL {}\ \BBA {} Koutrakis%
}{%
Lee%
\ \protect \BOthers {.}}{%
{\protect \APACyear {2011}}%
}]{%
lee2011novel}
\APACinsertmetastar {%
lee2011novel}%
\begin{APACrefauthors}%
Lee, H.%
, Liu, Y.%
, Coull, B.%
, Schwartz, J.%
\BCBL {} Koutrakis, P.%
\end{APACrefauthors}%
\unskip\
\newblock
\APACrefYearMonthDay{2011}{}{}.
\newblock
{\BBOQ}\APACrefatitle {A novel calibration approach of {MODIS AOD} data to
  predict {PM2.5} concentrations} {A novel calibration approach of {MODIS AOD}
  data to predict {PM2.5} concentrations}.{\BBCQ}
\newblock
\APACjournalVolNumPages{Atmospheric Chemistry and Physics}{11}{15}{7991--8002}.
\newblock

\newblock

\PrintBackRefs{\CurrentBib}

\bibitem [\protect \citeauthoryear {%
L.~Li%
}{%
L.~Li%
}{%
{\protect \APACyear {2020}}%
}]{%
rs12020264}
\APACinsertmetastar {%
rs12020264}%
\begin{APACrefauthors}%
Li, L.%
\end{APACrefauthors}%
\unskip\
\newblock
\APACrefYearMonthDay{2020}{}{}.
\newblock
{\BBOQ}\APACrefatitle {A Robust Deep Learning Approach for Spatiotemporal
  Estimation of Satellite AOD and {PM2.5}} {A robust deep learning approach for
  spatiotemporal estimation of satellite aod and {PM2.5}}.{\BBCQ}
\newblock
\APACjournalVolNumPages{Remote Sensing}{12}{2}{}.
\newblock

\newblock

\PrintBackRefs{\CurrentBib}

\bibitem [\protect \citeauthoryear {%
T.~Li%
, Shen%
, Yuan%
, Zhang%
\BCBL {}\ \BBA {} Zhang%
}{%
T.~Li%
\ \protect \BOthers {.}}{%
{\protect \APACyear {2017}}%
}]{%
https://doi.org/10.1002/2017GL075710}
\APACinsertmetastar {%
https://doi.org/10.1002/2017GL075710}%
\begin{APACrefauthors}%
Li, T.%
, Shen, H.%
, Yuan, Q.%
, Zhang, X.%
\BCBL {} Zhang, L.%
\end{APACrefauthors}%
\unskip\
\newblock
\APACrefYearMonthDay{2017}{}{}.
\newblock
{\BBOQ}\APACrefatitle {Estimating Ground-Level {PM2.5} by Fusing Satellite and
  Station Observations: A Geo-Intelligent Deep Learning Approach} {Estimating
  ground-level {PM2.5} by fusing satellite and station observations: A
  geo-intelligent deep learning approach}.{\BBCQ}
\newblock
\APACjournalVolNumPages{Geophysical Research Letters}{44}{23}{11,985-11,993}.
\newblock

\newblock

\PrintBackRefs{\CurrentBib}

\bibitem [\protect \citeauthoryear {%
Lin%
\ \protect \BOthers {.}}{%
Lin%
\ \protect \BOthers {.}}{%
{\protect \APACyear {2015}}%
}]{%
LIN2015117}
\APACinsertmetastar {%
LIN2015117}%
\begin{APACrefauthors}%
Lin, C.%
, Li, Y.%
, Yuan, Z.%
, Lau, A.K.%
, Li, C.%
\BCBL {} Fung, J.C.%
\end{APACrefauthors}%
\unskip\
\newblock
\APACrefYearMonthDay{2015}{}{}.
\newblock
{\BBOQ}\APACrefatitle {Using satellite remote sensing data to estimate the
  high-resolution distribution of ground-level {PM2.5}} {Using satellite remote
  sensing data to estimate the high-resolution distribution of ground-level
  {PM2.5}}.{\BBCQ}
\newblock
\APACjournalVolNumPages{Remote Sensing of Environment}{156}{}{117-128}.
\newblock

\newblock

\PrintBackRefs{\CurrentBib}

\bibitem [\protect \citeauthoryear {%
Lippmann%
, Ito%
, Nadas%
\BCBL {}\ \BBA {} Burnett%
}{%
Lippmann%
\ \protect \BOthers {.}}{%
{\protect \APACyear {2000}}%
}]{%
lippmann2000association}
\APACinsertmetastar {%
lippmann2000association}%
\begin{APACrefauthors}%
Lippmann, M.%
, Ito, K.%
, Nadas, A.%
\BCBL {} Burnett, R.%
\end{APACrefauthors}%
\unskip\
\newblock
\APACrefYearMonthDay{2000}{}{}.
\newblock
{\BBOQ}\APACrefatitle {Association of particulate matter components with daily
  mortality and morbidity in urban populations} {Association of particulate
  matter components with daily mortality and morbidity in urban
  populations}.{\BBCQ}
\newblock
\APACjournalVolNumPages{Research Report (Health Effects
  Institute)}{}{95}{5--72}.
\newblock

\newblock

\PrintBackRefs{\CurrentBib}

\bibitem [\protect \citeauthoryear {%
T.~Liu%
, Li%
, Yu%
\BCBL {}\ \BBA {} Qin%
}{%
T.~Liu%
\ \protect \BOthers {.}}{%
{\protect \APACyear {2017}}%
}]{%
liu2017nirs}
\APACinsertmetastar {%
liu2017nirs}%
\begin{APACrefauthors}%
Liu, T.%
, Li, Z.%
, Yu, C.%
\BCBL {} Qin, Y.%
\end{APACrefauthors}%
\unskip\
\newblock
\APACrefYearMonthDay{2017}{}{}.
\newblock
{\BBOQ}\APACrefatitle {{NIRS} feature extraction based on deep auto-encoder
  neural network} {{NIRS} feature extraction based on deep auto-encoder neural
  network}.{\BBCQ}
\newblock
\APACjournalVolNumPages{Infrared Physics \& Technology}{87}{}{124--128}.
\newblock

\newblock

\PrintBackRefs{\CurrentBib}

\bibitem [\protect \citeauthoryear {%
Y.~Liu%
\ \protect \BOthers {.}}{%
Y.~Liu%
\ \protect \BOthers {.}}{%
{\protect \APACyear {2004}}%
}]{%
https://doi.org/10.1029/2004JD005025}
\APACinsertmetastar {%
https://doi.org/10.1029/2004JD005025}%
\begin{APACrefauthors}%
Liu, Y.%
, Park, R.J.%
, Jacob, D.J.%
, Li, Q.%
, Kilaru, V.%
\BCBL {} Sarnat, J.A.%
\end{APACrefauthors}%
\unskip\
\newblock
\APACrefYearMonthDay{2004}{}{}.
\newblock
{\BBOQ}\APACrefatitle {Mapping annual mean ground-level {PM2.5} concentrations
  using Multiangle Imaging Spectroradiometer aerosol optical thickness over the
  contiguous {United States}} {Mapping annual mean ground-level {PM2.5}
  concentrations using multiangle imaging spectroradiometer aerosol optical
  thickness over the contiguous {United States}}.{\BBCQ}
\newblock
\APACjournalVolNumPages{Journal of Geophysical Research:
  Atmospheres}{109}{D22}{}.
\newblock

\newblock

\PrintBackRefs{\CurrentBib}

\bibitem [\protect \citeauthoryear {%
Lyapustin%
\ \BBA {} Wang%
}{%
Lyapustin%
\ \BBA {} Wang%
}{%
{\protect \APACyear {2018}}%
}]{%
lyapustin2018modis}
\APACinsertmetastar {%
lyapustin2018modis}%
\begin{APACrefauthors}%
Lyapustin, A.%
\BCBT {}\ \BBA {} Wang, Y.%
\end{APACrefauthors}%
\unskip\
\newblock
\APACrefYearMonthDay{2018}{}{}.
\newblock
{\BBOQ}\APACrefatitle {{MODIS} Multi-Angle Implementation of Atmospheric
  Correction ({MAIAC}) Data User’s Guide} {{MODIS} multi-angle implementation
  of atmospheric correction ({MAIAC}) data user’s guide}.{\BBCQ}
\newblock
\APACjournalVolNumPages{NASA: Greenbelt, MD, USA}{}{}{}.
\newblock

\newblock

\PrintBackRefs{\CurrentBib}

\bibitem [\protect \citeauthoryear {%
Lyapustin%
, Wang%
, Korkin%
\BCBL {}\ \BBA {} Huang%
}{%
Lyapustin%
\ \protect \BOthers {.}}{%
{\protect \APACyear {2018}}%
}]{%
amt-11-5741-2018}
\APACinsertmetastar {%
amt-11-5741-2018}%
\begin{APACrefauthors}%
Lyapustin, A.%
, Wang, Y.%
, Korkin, S.%
\BCBL {} Huang, D.%
\end{APACrefauthors}%
\unskip\
\newblock
\APACrefYearMonthDay{2018}{}{}.
\newblock
{\BBOQ}\APACrefatitle {{MODIS} Collection 6 {MAIAC} algorithm} {{MODIS}
  collection 6 {MAIAC} algorithm}.{\BBCQ}
\newblock
\APACjournalVolNumPages{Atmospheric Measurement
  Techniques}{11}{10}{5741--5765}.
\newblock

\newblock

\PrintBackRefs{\CurrentBib}

\bibitem [\protect \citeauthoryear {%
Lyapustin%
, Zhao%
\BCBL {}\ \BBA {} Wang%
}{%
Lyapustin%
\ \protect \BOthers {.}}{%
{\protect \APACyear {2021}}%
}]{%
10.3389/frsen.2021.712093}
\APACinsertmetastar {%
10.3389/frsen.2021.712093}%
\begin{APACrefauthors}%
Lyapustin, A.%
, Zhao, F.%
\BCBL {} Wang, Y.%
\end{APACrefauthors}%
\unskip\
\newblock
\APACrefYearMonthDay{2021}{}{}.
\newblock
{\BBOQ}\APACrefatitle {A Comparison of Multi-Angle Implementation of
  Atmospheric Correction and {MOD09} Daily Surface Reflectance Products From
  {MODIS}} {A comparison of multi-angle implementation of atmospheric
  correction and {MOD09} daily surface reflectance products from
  {MODIS}}.{\BBCQ}
\newblock
\APACjournalVolNumPages{Frontiers in Remote Sensing}{2}{}{}.
\newblock

\newblock

\PrintBackRefs{\CurrentBib}

\bibitem [\protect \citeauthoryear {%
Ma%
\ \protect \BOthers {.}}{%
Ma%
\ \protect \BOthers {.}}{%
{\protect \APACyear {2016}}%
}]{%
ma2016satellite}
\APACinsertmetastar {%
ma2016satellite}%
\begin{APACrefauthors}%
Ma, Z.%
, Hu, X.%
, Sayer, A.M.%
, Levy, R.%
, Zhang, Q.%
, Xue, Y.%
\BDBL {}Liu, Y.%
\end{APACrefauthors}%
\unskip\
\newblock
\APACrefYearMonthDay{2016}{}{}.
\newblock
{\BBOQ}\APACrefatitle {Satellite-based spatiotemporal trends in {PM2.5}
  concentrations: {China}, 2004--2013} {Satellite-based spatiotemporal trends
  in {PM2.5} concentrations: {China}, 2004--2013}.{\BBCQ}
\newblock
\APACjournalVolNumPages{Environmental health perspectives}{124}{2}{184--192}.
\newblock

\newblock

\PrintBackRefs{\CurrentBib}

\bibitem [\protect \citeauthoryear {%
Nabavi%
, Haimberger%
\BCBL {}\ \BBA {} Abbasi%
}{%
Nabavi%
\ \protect \BOthers {.}}{%
{\protect \APACyear {2019}}%
}]{%
NABAVI2019889}
\APACinsertmetastar {%
NABAVI2019889}%
\begin{APACrefauthors}%
Nabavi, S.O.%
, Haimberger, L.%
\BCBL {} Abbasi, E.%
\end{APACrefauthors}%
\unskip\
\newblock
\APACrefYearMonthDay{2019}{}{}.
\newblock
{\BBOQ}\APACrefatitle {Assessing {PM2.5} concentrations in {Tehran}, {Iran},
  from space using {MAIAC}, deep blue, and dark target {AOD} and machine
  learning algorithms} {Assessing {PM2.5} concentrations in {Tehran}, {Iran},
  from space using {MAIAC}, deep blue, and dark target {AOD} and machine
  learning algorithms}.{\BBCQ}
\newblock
\APACjournalVolNumPages{Atmospheric Pollution Research}{10}{3}{889-903}.
\newblock

\newblock

\PrintBackRefs{\CurrentBib}

\bibitem [\protect \citeauthoryear {%
Ni%
, Cao%
, Zhou%
, Cui%
\BCBL {}\ \BBA {} P.~Singh%
}{%
Ni%
\ \protect \BOthers {.}}{%
{\protect \APACyear {2018}}%
}]{%
atmos9030105}
\APACinsertmetastar {%
atmos9030105}%
\begin{APACrefauthors}%
Ni, X.%
, Cao, C.%
, Zhou, Y.%
, Cui, X.%
\BCBL {} P.~Singh, R.%
\end{APACrefauthors}%
\unskip\
\newblock
\APACrefYearMonthDay{2018}{}{}.
\newblock
{\BBOQ}\APACrefatitle {Spatio-Temporal Pattern Estimation of {PM2.5} in
  {Beijing-Tianjin-Hebei} Region Based on {MODIS AOD} and Meteorological Data
  Using the Back Propagation Neural Network} {Spatio-temporal pattern
  estimation of {PM2.5} in {Beijing-Tianjin-Hebei} region based on {MODIS AOD}
  and meteorological data using the back propagation neural network}.{\BBCQ}
\newblock
\APACjournalVolNumPages{Atmosphere}{9}{3}{}.
\newblock

\newblock

\PrintBackRefs{\CurrentBib}

\bibitem [\protect \citeauthoryear {%
Sayer%
, Hsu%
, Bettenhausen%
, Jeong%
\BCBL {}\ \BBA {} Meister%
}{%
Sayer%
\ \protect \BOthers {.}}{%
{\protect \APACyear {2015}}%
}]{%
sayer2015effect}
\APACinsertmetastar {%
sayer2015effect}%
\begin{APACrefauthors}%
Sayer, A.%
, Hsu, N.%
, Bettenhausen, C.%
, Jeong, M\BHBI J.%
\BCBL {} Meister, G.%
\end{APACrefauthors}%
\unskip\
\newblock
\APACrefYearMonthDay{2015}{}{}.
\newblock
{\BBOQ}\APACrefatitle {Effect of {MODIS Terra} radiometric calibration
  improvements on Collection 6 Deep Blue aerosol products: Validation and
  {Terra/Aqua} consistency} {Effect of {MODIS Terra} radiometric calibration
  improvements on collection 6 deep blue aerosol products: Validation and
  {Terra/Aqua} consistency}.{\BBCQ}
\newblock
\APACjournalVolNumPages{Journal of Geophysical Research:
  Atmospheres}{120}{23}{12--157}.
\newblock

\newblock

\PrintBackRefs{\CurrentBib}

\bibitem [\protect \citeauthoryear {%
Song%
, Jia%
, Huang%
\BCBL {}\ \BBA {} Zhang%
}{%
Song%
\ \protect \BOthers {.}}{%
{\protect \APACyear {2014}}%
}]{%
SONG20141}
\APACinsertmetastar {%
SONG20141}%
\begin{APACrefauthors}%
Song, W.%
, Jia, H.%
, Huang, J.%
\BCBL {} Zhang, Y.%
\end{APACrefauthors}%
\unskip\
\newblock
\APACrefYearMonthDay{2014}{}{}.
\newblock
{\BBOQ}\APACrefatitle {A satellite-based geographically weighted regression
  model for regional {PM2.5} estimation over the {Pearl River Delta} region in
  {China}} {A satellite-based geographically weighted regression model for
  regional {PM2.5} estimation over the {Pearl River Delta} region in
  {China}}.{\BBCQ}
\newblock
\APACjournalVolNumPages{Remote Sensing of Environment}{154}{}{1-7}.
\newblock

\newblock

\PrintBackRefs{\CurrentBib}

\bibitem [\protect \citeauthoryear {%
Sotoudeheian%
\ \BBA {} Arhami%
}{%
Sotoudeheian%
\ \BBA {} Arhami%
}{%
{\protect \APACyear {2014}}%
}]{%
sotoudeheian2014estimating}
\APACinsertmetastar {%
sotoudeheian2014estimating}%
\begin{APACrefauthors}%
Sotoudeheian, S.%
\BCBT {}\ \BBA {} Arhami, M.%
\end{APACrefauthors}%
\unskip\
\newblock
\APACrefYearMonthDay{2014}{}{}.
\newblock
{\BBOQ}\APACrefatitle {Estimating ground-level {PM10} using satellite remote
  sensing and ground-based meteorological measurements over {Tehran}}
  {Estimating ground-level {PM10} using satellite remote sensing and
  ground-based meteorological measurements over {Tehran}}.{\BBCQ}
\newblock
\APACjournalVolNumPages{Journal of Environmental Health Science and
  Engineering}{12}{1}{1--13}.
\newblock

\newblock

\PrintBackRefs{\CurrentBib}

\bibitem [\protect \citeauthoryear {%
Sun%
, Gong%
\BCBL {}\ \BBA {} Zhou%
}{%
Sun%
\ \protect \BOthers {.}}{%
{\protect \APACyear {2021}}%
}]{%
SUN2021144502}
\APACinsertmetastar {%
SUN2021144502}%
\begin{APACrefauthors}%
Sun, J.%
, Gong, J.%
\BCBL {} Zhou, J.%
\end{APACrefauthors}%
\unskip\
\newblock
\APACrefYearMonthDay{2021}{}{}.
\newblock
{\BBOQ}\APACrefatitle {Estimating hourly {PM2.5} concentrations in {Beijing}
  with satellite aerosol optical depth and a random forest approach}
  {Estimating hourly {PM2.5} concentrations in {Beijing} with satellite aerosol
  optical depth and a random forest approach}.{\BBCQ}
\newblock
\APACjournalVolNumPages{Science of The Total Environment}{762}{}{144502}.
\newblock

\newblock

\PrintBackRefs{\CurrentBib}

\bibitem [\protect \citeauthoryear {%
Van~Donkelaar%
\ \protect \BOthers {.}}{%
Van~Donkelaar%
\ \protect \BOthers {.}}{%
{\protect \APACyear {2010}}%
}]{%
van2010global}
\APACinsertmetastar {%
van2010global}%
\begin{APACrefauthors}%
Van~Donkelaar, A.%
, Martin, R.V.%
, Brauer, M.%
, Kahn, R.%
, Levy, R.%
, Verduzco, C.%
\BCBL {} Villeneuve, P.J.%
\end{APACrefauthors}%
\unskip\
\newblock
\APACrefYearMonthDay{2010}{}{}.
\newblock
{\BBOQ}\APACrefatitle {Global estimates of ambient fine particulate matter
  concentrations from satellite-based aerosol optical depth: development and
  application} {Global estimates of ambient fine particulate matter
  concentrations from satellite-based aerosol optical depth: development and
  application}.{\BBCQ}
\newblock
\APACjournalVolNumPages{Environmental health perspectives}{118}{6}{847--855}.
\newblock

\newblock

\PrintBackRefs{\CurrentBib}

\bibitem [\protect \citeauthoryear {%
J.~Wang%
\ \BBA {} Christopher%
}{%
J.~Wang%
\ \BBA {} Christopher%
}{%
{\protect \APACyear {2003}}%
{\protect \APACexlab {{\protect \BCnt {1}}}}}]{%
christopher}
\APACinsertmetastar {%
christopher}%
\begin{APACrefauthors}%
Wang, J.%
\BCBT {}\ \BBA {} Christopher, S.A.%
\end{APACrefauthors}%
\unskip\
\newblock
\APACrefYearMonthDay{2003{\protect \BCnt {1}}}{}{}.
\newblock
{\BBOQ}\APACrefatitle {Intercomparison between satellite-derived aerosol
  optical thickness and {PM2.5} mass: Implications for air quality studies}
  {Intercomparison between satellite-derived aerosol optical thickness and
  {PM2.5} mass: Implications for air quality studies}.{\BBCQ}
\newblock
\APACjournalVolNumPages{Geophysical Research Letters}{30}{21}{}.
\newblock

\newblock

\PrintBackRefs{\CurrentBib}

\bibitem [\protect \citeauthoryear {%
J.~Wang%
\ \BBA {} Christopher%
}{%
J.~Wang%
\ \BBA {} Christopher%
}{%
{\protect \APACyear {2003}}%
{\protect \APACexlab {{\protect \BCnt {2}}}}}]{%
https://doi.org/10.1029/2003GL018174}
\APACinsertmetastar {%
https://doi.org/10.1029/2003GL018174}%
\begin{APACrefauthors}%
Wang, J.%
\BCBT {}\ \BBA {} Christopher, S.A.%
\end{APACrefauthors}%
\unskip\
\newblock
\APACrefYearMonthDay{2003{\protect \BCnt {2}}}{}{}.
\newblock
{\BBOQ}\APACrefatitle {Intercomparison between satellite-derived aerosol
  optical thickness and {PM2.5} mass: Implications for air quality studies}
  {Intercomparison between satellite-derived aerosol optical thickness and
  {PM2.5} mass: Implications for air quality studies}.{\BBCQ}
\newblock
\APACjournalVolNumPages{Geophysical Research Letters}{30}{21}{}.
\newblock

\newblock

\PrintBackRefs{\CurrentBib}

\bibitem [\protect \citeauthoryear {%
X.~Wang%
\ \BBA {} Sun%
}{%
X.~Wang%
\ \BBA {} Sun%
}{%
{\protect \APACyear {2019}}%
}]{%
WANG2019128}
\APACinsertmetastar {%
WANG2019128}%
\begin{APACrefauthors}%
Wang, X.%
\BCBT {}\ \BBA {} Sun, W.%
\end{APACrefauthors}%
\unskip\
\newblock
\APACrefYearMonthDay{2019}{}{}.
\newblock
{\BBOQ}\APACrefatitle {Meteorological parameters and gaseous pollutant
  concentrations as predictors of daily continuous {PM2.5} concentrations using
  deep neural network in {Beijing–Tianjin–Hebei, China}} {Meteorological
  parameters and gaseous pollutant concentrations as predictors of daily
  continuous {PM2.5} concentrations using deep neural network in
  {Beijing–Tianjin–Hebei, China}}.{\BBCQ}
\newblock
\APACjournalVolNumPages{Atmospheric Environment}{211}{}{128-137}.
\newblock

\newblock

\PrintBackRefs{\CurrentBib}

\bibitem [\protect \citeauthoryear {%
Z.~Wang%
, Chen%
, Tao%
, Zhang%
\BCBL {}\ \BBA {} Su%
}{%
Z.~Wang%
\ \protect \BOthers {.}}{%
{\protect \APACyear {2010}}%
}]{%
wang2010satellite}
\APACinsertmetastar {%
wang2010satellite}%
\begin{APACrefauthors}%
Wang, Z.%
, Chen, L.%
, Tao, J.%
, Zhang, Y.%
\BCBL {} Su, L.%
\end{APACrefauthors}%
\unskip\
\newblock
\APACrefYearMonthDay{2010}{}{}.
\newblock
{\BBOQ}\APACrefatitle {Satellite-based estimation of regional particulate
  matter ({PM}) in {Beijing} using vertical-and-{RH} correcting method}
  {Satellite-based estimation of regional particulate matter ({PM}) in
  {Beijing} using vertical-and-{RH} correcting method}.{\BBCQ}
\newblock
\APACjournalVolNumPages{Remote sensing of environment}{114}{1}{50--63}.
\newblock

\newblock

\PrintBackRefs{\CurrentBib}

\bibitem [\protect \citeauthoryear {%
Weizhen%
\ \protect \BOthers {.}}{%
Weizhen%
\ \protect \BOthers {.}}{%
{\protect \APACyear {2014}}%
}]{%
Weizhen_2014}
\APACinsertmetastar {%
Weizhen_2014}%
\begin{APACrefauthors}%
Weizhen, H.%
, Zhengqiang, L.%
, Yuhuan, Z.%
, Hua, X.%
, Ying, Z.%
, Kaitao, L.%
\BDBL {}Yan, M.%
\end{APACrefauthors}%
\unskip\
\newblock
\APACrefYearMonthDay{2014}{}{}.
\newblock
{\BBOQ}\APACrefatitle {Using support vector regression to predict {PM}10 and
  {PM}2.5} {Using support vector regression to predict {PM}10 and
  {PM}2.5}.{\BBCQ}
\newblock
\APACjournalVolNumPages{{IOP} Conference Series: Earth and Environmental
  Science}{17}{}{012268}.
\newblock

\newblock

\PrintBackRefs{\CurrentBib}

\bibitem [\protect \citeauthoryear {%
Xiao%
\ \protect \BOthers {.}}{%
Xiao%
\ \protect \BOthers {.}}{%
{\protect \APACyear {2017}}%
}]{%
xiao2017full}
\APACinsertmetastar {%
xiao2017full}%
\begin{APACrefauthors}%
Xiao, Q.%
, Wang, Y.%
, Chang, H.H.%
, Meng, X.%
, Geng, G.%
, Lyapustin, A.%
\BCBL {} Liu, Y.%
\end{APACrefauthors}%
\unskip\
\newblock
\APACrefYearMonthDay{2017}{}{}.
\newblock
{\BBOQ}\APACrefatitle {Full-coverage high-resolution daily {PM2.5} estimation
  using {MAIAC AOD} in the {Yangtze} River Delta of {China}} {Full-coverage
  high-resolution daily {PM2.5} estimation using {MAIAC AOD} in the {Yangtze}
  river delta of {China}}.{\BBCQ}
\newblock
\APACjournalVolNumPages{Remote Sensing of Environment}{199}{}{437--446}.
\newblock

\newblock

\PrintBackRefs{\CurrentBib}

\bibitem [\protect \citeauthoryear {%
Yang%
, Rahardja%
\BCBL {}\ \BBA {} Fr{\"a}nti%
}{%
Yang%
\ \protect \BOthers {.}}{%
{\protect \APACyear {2019}}%
}]{%
yang2019outlier}
\APACinsertmetastar {%
yang2019outlier}%
\begin{APACrefauthors}%
Yang, J.%
, Rahardja, S.%
\BCBL {} Fr{\"a}nti, P.%
\end{APACrefauthors}%
\unskip\
\newblock
\APACrefYearMonthDay{2019}{}{}.
\newblock
{\BBOQ}\APACrefatitle {Outlier detection: how to threshold outlier scores?}
  {Outlier detection: how to threshold outlier scores?}{\BBCQ}
\newblock
 \APACrefbtitle {Proceedings of the international conference on artificial
  intelligence, information processing and cloud computing} {Proceedings of the
  international conference on artificial intelligence, information processing
  and cloud computing}\ (\BPGS\ 1--6).
\PrintBackRefs{\CurrentBib}

\bibitem [\protect \citeauthoryear {%
Yao%
, Si%
, Li%
\BCBL {}\ \BBA {} Wu%
}{%
Yao%
\ \protect \BOthers {.}}{%
{\protect \APACyear {2018}}%
}]{%
YAO2018819}
\APACinsertmetastar {%
YAO2018819}%
\begin{APACrefauthors}%
Yao, F.%
, Si, M.%
, Li, W.%
\BCBL {} Wu, J.%
\end{APACrefauthors}%
\unskip\
\newblock
\APACrefYearMonthDay{2018}{}{}.
\newblock
{\BBOQ}\APACrefatitle {A multidimensional comparison between {MODIS} and {VIIRS
  AOD} in estimating ground-level {PM2.5} concentrations over a heavily
  polluted region in {China}} {A multidimensional comparison between {MODIS}
  and {VIIRS AOD} in estimating ground-level {PM2.5} concentrations over a
  heavily polluted region in {China}}.{\BBCQ}
\newblock
\APACjournalVolNumPages{Science of The Total Environment}{618}{}{819-828}.
\newblock

\newblock

\PrintBackRefs{\CurrentBib}

\bibitem [\protect \citeauthoryear {%
You%
, Zang%
, Pan%
, Zhang%
\BCBL {}\ \BBA {} Chen%
}{%
You%
\ \protect \BOthers {.}}{%
{\protect \APACyear {2015}}%
}]{%
YOU20151156}
\APACinsertmetastar {%
YOU20151156}%
\begin{APACrefauthors}%
You, W.%
, Zang, Z.%
, Pan, X.%
, Zhang, L.%
\BCBL {} Chen, D.%
\end{APACrefauthors}%
\unskip\
\newblock
\APACrefYearMonthDay{2015}{}{}.
\newblock
{\BBOQ}\APACrefatitle {Estimating {PM2.5} in {Xi'an, China} using aerosol
  optical depth: A comparison between the {MODIS} and {MISR} retrieval models}
  {Estimating {PM2.5} in {Xi'an, China} using aerosol optical depth: A
  comparison between the {MODIS} and {MISR} retrieval models}.{\BBCQ}
\newblock
\APACjournalVolNumPages{Science of The Total Environment}{505}{}{1156-1165}.
\newblock

\newblock

\PrintBackRefs{\CurrentBib}

\bibitem [\protect \citeauthoryear {%
Zamani~Joharestani%
, Cao%
, Ni%
, Bashir%
\BCBL {}\ \BBA {} Talebiesfandarani%
}{%
Zamani~Joharestani%
\ \protect \BOthers {.}}{%
{\protect \APACyear {2019}}%
}]{%
atmos10070373}
\APACinsertmetastar {%
atmos10070373}%
\begin{APACrefauthors}%
Zamani~Joharestani, M.%
, Cao, C.%
, Ni, X.%
, Bashir, B.%
\BCBL {} Talebiesfandarani, S.%
\end{APACrefauthors}%
\unskip\
\newblock
\APACrefYearMonthDay{2019}{}{}.
\newblock
{\BBOQ}\APACrefatitle {{PM2.5} Prediction Based on Random Forest, {XGBoost},
  and Deep Learning Using Multisource Remote Sensing Data} {{PM2.5} prediction
  based on random forest, {XGBoost}, and deep learning using multisource remote
  sensing data}.{\BBCQ}
\newblock
\APACjournalVolNumPages{Atmosphere}{10}{7}{}.
\newblock

\newblock

\PrintBackRefs{\CurrentBib}

\bibitem [\protect \citeauthoryear {%
Zhou%
\ \BBA {} Feng%
}{%
Zhou%
\ \BBA {} Feng%
}{%
{\protect \APACyear {2017}}%
}]{%
ijcai2017-497}
\APACinsertmetastar {%
ijcai2017-497}%
\begin{APACrefauthors}%
Zhou, Z\BHBI H.%
\BCBT {}\ \BBA {} Feng, J.%
\end{APACrefauthors}%
\unskip\
\newblock
\APACrefYearMonthDay{2017}{}{}.
\newblock
{\BBOQ}\APACrefatitle {Deep Forest: Towards An Alternative to Deep Neural
  Networks} {Deep forest: Towards an alternative to deep neural
  networks}.{\BBCQ}
\newblock
 \APACrefbtitle {Proceedings of the Twenty-Sixth International Joint Conference
  on Artificial Intelligence, {IJCAI-17}} {Proceedings of the twenty-sixth
  international joint conference on artificial intelligence, {IJCAI-17}}\
  (\BPGS\ 3553--3559).
\PrintBackRefs{\CurrentBib}

\end{thebibliography}


\end{document}